\pdfoutput=1
\documentclass{article}
\usepackage{iclr2027_conference,times}

\usepackage{amsmath,amsfonts,bm}

\def\eqref#1{equation~\ref{#1}}

\def\1{\bm{1}}

\DeclareMathAlphabet{\mathsfit}{\encodingdefault}{\sfdefault}{m}{sl}
\SetMathAlphabet{\mathsfit}{bold}{\encodingdefault}{\sfdefault}{bx}{n}

\usepackage{hyperref}
\usepackage{url}
\usepackage{booktabs}
\usepackage{graphicx}
\usepackage{float}
\usepackage{wrapfig}
\makeatletter\newcommand{\captionof}[2]{\def\@captype{#1}\caption{#2}}\makeatother
\usepackage{amsmath}
\usepackage{amssymb}
\usepackage{tikz}
\usepackage{paralist}
\usetikzlibrary{positioning,arrows.meta,calc,decorations.pathreplacing}

\title{CLASP: Continual Low-rank Adapters \\ for Spatially Placed Concepts \\ from One Hypernetwork}

\author{Wojciech Gromski$^{1,2}$\quad Patryk Krukowski$^{3,4}$\quad Jan Miksa$^{2,3}$ \\ \bfseries Maciej Zieba$^{1,5}$\quad Przemys{\l}aw Spurek$^{2,3}$ \\ {\normalfont $^{1}$Wroc{\l}aw University of Science and Technology\quad $^{2}$IDEAS Research Institute} \\ {\normalfont $^{3}$Jagiellonian University, Krak\'ow\quad $^{4}$AKCES NCBR\quad $^{5}$Tooploox} \\[4pt] {\normalfont\small\hypersetup{pdfborder={0 0 0}}\href{https://github.com/genwro-ai/clasp}{\textcolor{claspblue}{\faGithub~\texttt{github.com/genwro-ai/clasp}}}\qquad\href{https://genwro-ai.github.io/clasp}{\textcolor{claspblue}{\faGlobe~\texttt{genwro-ai.github.io/clasp}}}}}

\usepackage{xcolor}
\usepackage{fontawesome5}
\definecolor{claspblue}{RGB}{56,117,186}
\newif\ifdraftnotes
\draftnotesfalse

\newcommand{\hl}{\smash{h^{(\ell)}_\phi}}
\newcommand{\htok}{\smash{h^{\mathrm{tok}}_\phi}}
\newcommand{\hfilm}{\smash{h^{\mathrm{FiLM}}_\phi}}

\iclrfinalcopy

\begin{document}

\maketitle
\vspace{-14pt}
\begin{figure}[H]
  \centering\setlength{\abovecaptionskip}{4pt}
  \includegraphics[width=\linewidth]{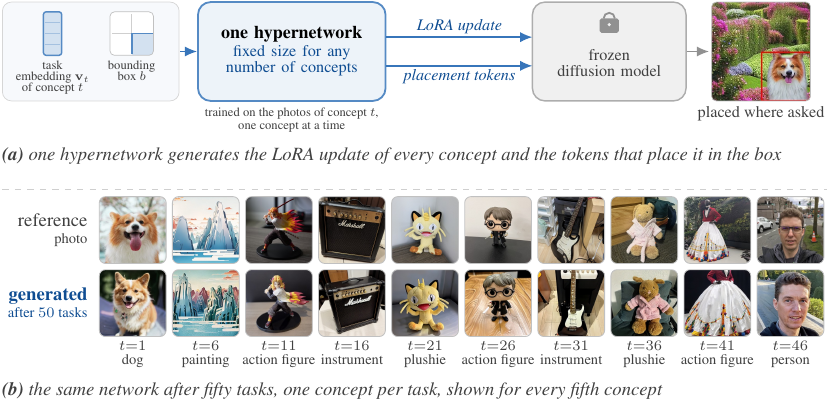}
  \vspace{-7pt}
  \caption{
    \textbf{(a)} A fixed-size hypernetwork takes the task embedding of a concept and,
    optionally, a bounding box, and generates the parameters needed to personalize
    the frozen diffusion model: a low-rank update for its cross-attention layers,
    which captures \emph{what} the concept looks like, and placement tokens, which
    control \emph{where} it appears. The resulting model generates the personalized
    concept at the requested location while following the input prompt.
    \textbf{(b)} Every fifth concept from a fifty-concept sequence. Each column shows
    a reference image (top) and an image generated by the final model after all
    fifty concepts have been learned sequentially (bottom).
    }
  \label{fig:teaser}
\end{figure}
\vspace{-12pt}
\begin{abstract}
Continual personalization of text-to-image diffusion models requires sequentially acquiring new concepts while retaining previously learned ones. However, existing methods either suffer from catastrophic forgetting or rely on storing additional concept-specific parameters and spatial components, causing their parameter footprint to grow with the concept stream. This limits their ability to scale to long sequences of personalization tasks. We propose a rehearsal-free approach that uses a single fixed-size hypernetwork to continually personalize a frozen diffusion model. Instead of expanding the model as new concepts are acquired, the hypernetwork dynamically produces the concept-specific adaptations required for personalization while preserving previously learned concepts. Our framework further integrates spatial control into the personalization process, allowing users to specify where a personalized concept should appear without introducing additional per-concept components. This formulation enables continual personalization with a parameter footprint that remains independent of the number of learned concepts, aside from compact concept representations. Experiments demonstrate strong retention of previously learned concepts and reliable spatial grounding, matching or improving upon existing methods while scaling effectively to long streams of personalization tasks.
\end{abstract}
\vspace{-18pt}

\section{Introduction}
\label{sec:intro}

Continual learning~\citep{mccloskey1989catastrophic,wang2024continualsurvey} considers settings in which models must acquire new knowledge while retaining what they have previously learned. This problem arises in \emph{text-to-image diffusion models}~\citep{Ho2020DenoisingDP,rombach2022ldm}, where users may progressively personalize a pretrained model to new visual concepts. Personalization adapts a model to a specific subject or style from a small set of reference images~\citep{gal2023textualinversion,ruiz2023dreambooth}. The model can then generate that concept under new prompts. When concepts arrive sequentially, the model must acquire each new concept without forgetting earlier ones. This is \emph{continual personalization}. This setting is harder when storing previous training data is undesirable or infeasible~\citep{carlini2023extractingtrainingdatadiffusion,Dubinski_2025_CVPR,Verwimp2023ContinualLA}.

Existing continual personalization methods typically store a separate adaptation for each concept~\citep{dong2024continually}, so the number of stored parameters grows linearly with the number of concepts. This limits their scalability to long concept sequences. We address this limitation with a fixed-size \emph{hypernetwork}~\citep{Ha2017HyperNetworks}, which generates low-rank updates~\citep{hu2022lora} for a frozen diffusion model from a task embedding per concept, kept orthogonal to those of earlier concepts. Thus, the same network can generate adaptations for an increasing number of concepts, with each concept requiring only a compact representation rather than a separate adapter. Because the hypernetwork is updated continually, it can also forget previously learned adaptations. We mitigate this interference with output-space regularization~\citep{vonoswald2020hypernet}.

Personalization also requires controlling \emph{where} a concept appears. Existing concept adapters do not encode spatial information, while storing a separate spatial module per concept would again sacrifice scalability. We extend the same hypernetwork to generate placement tokens, learned from synthetic composites with controlled object locations. At inference time, a user-provided bounding box guides the personalized concept to the desired region without additional per-concept parameters.

We evaluate our approach on the \emph{Concept-Incremental Flexible Customization (CIFC)} benchmark~\citep{dong2024continually} and extend it to sequences of up to 100 concepts. On SD-1.5, we outperform CIDM~\citep{dong2024continually} in image alignment at matched text alignment 
and achieve approximately $4\times$ lower forgetting than CIDM and $23\times$ lower forgetting than fine-tuning. Along the stream, our method retains the first ten concepts better on average than CIDM at every length CIDM can reach, and it continues up to 100 concepts. We also demonstrate reliable placement, placing seven personalized concepts in their requested quadrants in every evaluated case.

The main contributions of this paper are summarized as follows:
\begin{compactitem}
    \item We propose a continual personalization method in which a fixed-size hypernetwork generates concept-specific low-rank adaptations without storing an adapter per concept.
    
    \item We extend the same hypernetwork to generate spatial conditioning for user-specified bounding boxes without additional per-concept modules.
    
    \item We evaluate our approach on sequences of up to 100 concepts, beyond the 37 that CIDM supports, and show that it outperforms CIDM in image alignment on SD-1.5 and achieves lower forgetting, while maintaining a fixed parameter count.
\end{compactitem}

\section{Related Work}
\label{sec:related}

\textbf{Continual learning and personalization.}
Continual learning trains a model on a sequence of tasks without forgetting earlier ones~\citep{mccloskey1989catastrophic,de2021continual,wang2024continualsurvey}, through regularization~\citep{kirkpatrick2017ewc,li2018lwf}, rehearsal~\citep{rebuffi2017icarl,shin2017continual} or parameter-efficient adaptation~\citep{wang2022dualprompt,smith2023codap,liang2024inflora}. Diffusion personalization adapts a text-to-image model to a subject or style from a few images, with learned tokens~\citep{gal2023textualinversion}, fine-tuning~\citep{ruiz2023dreambooth,kumari2023customdiffusion} or low-rank adapters~\citep{hu2022lora}. Continual personalization learns such concepts in sequence, through concept-specific isolation~\citep{smith2023continual,smith2024continual}, replay or distillation~\citep{sun2024create,jha2025mining}, or parameter merging and allocation~\citep{staniszewski2024low,liao2025continual}. CIDM~\citep{dong2024continually,dong2026crafting}, whose CIFC benchmark we adopt, keeps a separate set of low-rank parameters per concept, so its storage grows with the stream. CNS~\citep{liao2025continual} instead allocates part of the backbone to each concept, which bounds their number by its capacity. We keep the backbone fixed and generate every adaptation with one hypernetwork.

\textbf{Hypernetworks and concept separation.}
Hypernetworks~\citep{Ha2017HyperNetworks} generate the parameters of a target network from a compact task representation. In continual learning, \citet{vonoswald2020hypernet} regularize the generated parameters rather than the weights of the hypernetwork, and we adopt this output-space view for diffusion personalization. Hypernetworks have also been used for fast personalization~\citep{ruiz2023hyperdreambooth} and concept removal~\citep{wojcik2026unhype}, but not for learning concepts in sequence. Orthogonal Adaptation~\citep{po2024orthogonal}, SeqLoRA~\citep{parsa2026seqlora} and InfLoRA~\citep{liang2024inflora} instead place the adapters of different concepts in distinct subspaces, so their capacity stays tied to the adaptation parameters. Our network is shared by all concepts, and only the task embedding tells them apart. Appendix~\ref{app:related} discusses both lines of work in more detail.

\textbf{Spatial grounding.}
Controlling the spatial location of generated content has been studied through box-conditioned tokens~\citep{li2023gligen,jang2024reground}, attention manipulation~\citep{kim2023densediffusion}, and regional denoising~\citep{bartal2023multidiffusion}. These methods primarily provide spatial control for concepts specified directly by the text prompt, rather than for personalized subjects learned from reference images. Mix-of-Show~\citep{gu2023mixofshow} extends spatial control to personalized subjects, but relies on retaining a collection of concept-specific adapters. In our setting, adding spatial control should not reintroduce the per-concept parameter growth that limits existing methods. We generate both the concept adaptation and its spatial conditioning from the same compact concept representation, so a user-given box controls placement without any per-concept module.

\section{Preliminaries}
\label{sec:prelim}

\textbf{Latent diffusion models.}
A latent diffusion model~\citep{rombach2022ldm} generates an image by denoising a latent in the
space of a pretrained autoencoder. A UNet~\citep{ronneberger2015unet} $\boldsymbol{\varepsilon}_\theta$
predicts the noise in a noisy latent $\mathbf{x}_s$ at timestep $s$, conditioned on text embeddings
$\mathbf{c}$, and is trained with the loss $\mathcal{L}_{\mathrm{LDM}}$ of Eq.~\ref{eq:ldm}
(Appendix~\ref{app:ldm}). We reserve $t$ for the concept index. The text conditions the UNet through
cross-attention, whose query, key, value and output projections are $\mathbf{W}^Q$, $\mathbf{W}^K$,
$\mathbf{W}^V$ and $\mathbf{W}^O$, and we adapt all four.

\textbf{Low-rank adaptation.}
LoRA~\citep{hu2022lora} adapts a pretrained weight matrix
$\mathbf{W}_0\in\mathbb{R}^{m\times n}$ by adding a rank-$r$ update,
\begin{equation}
\mathbf{W}
=
\mathbf{W}_0 + \frac{\gamma}{r}\mathbf{A}\mathbf{B},
\qquad
\mathbf{A}\in\mathbb{R}^{m\times r},
\quad
\mathbf{B}\in\mathbb{R}^{r\times n},
\quad
r\ll\min(m,n),
\label{eq:lora}
\end{equation}
where $\gamma$ controls the update magnitude. Only $\mathbf{A}$ and $\mathbf{B}$
are trained, reducing the number of trainable parameters from $mn$ to $r(m+n)$.
Because our updates act only at selected token positions, they are applied as a
separate branch rather than merged into $\mathbf{W}_0$.

\textbf{Hypernetworks.}
A hypernetwork~\citep{Ha2017HyperNetworks} generates the parameters of another
network from a conditioning vector. We use a hypernetwork $h_\phi$ with parameters $\phi$,
shared across all concepts, that maps a task embedding $\mathbf{v}_t\in\mathbb{R}^{d_v}$ of concept
$t$ to the matrices $(\mathbf{A}_{t,\ell},\mathbf{B}_{t,\ell})$ of Eq.~\ref{eq:lora} for each of the
$L$ adapted UNet projections. The updates of individual concepts therefore need not be stored. Each concept
is represented by its task embedding $\mathbf{v}_t$, which generates its
adaptation on demand. When the
hypernetwork is trained sequentially, however, updating $\phi$ for a new
concept can alter the adaptations generated for earlier concepts, which our
regularizer addresses (Section~\ref{sec:content}).

\textbf{Problem setting.}
We consider a sequence of $T$ concepts indexed by $t=1,\ldots,T$, where each
concept $t$ is associated with a small reference set $\mathcal{D}_t$. Training
proceeds sequentially, with access only to the reference set of the current
concept. After learning concept $t$, the model should be able to generate any
previously learned concept $j\leq t$ in response to a natural-language prompt
specifying that concept, while preserving its visual identity as subsequent
concepts are learned. We additionally consider spatially controlled
generation, where a bounding box $\mathbf{b}\in[0,1]^4$ specifies the desired
location of the personalized concept in the generated image. Concept identity
is assumed to be known at generation time.

\section{Method}
\label{sec:method}

\begin{figure}[t]
  \centering
  \input{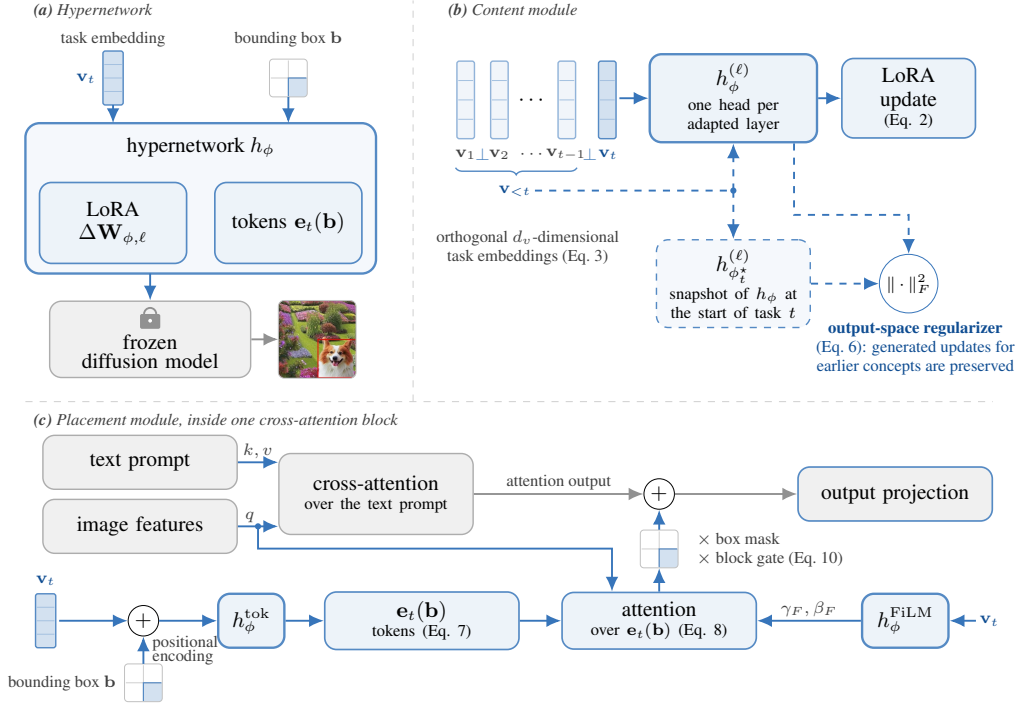}
  \caption{\textbf{Method overview.}
    \textbf{(a)} A hypernetwork takes the task embedding $\mathbf{v}_t$ and
    bounding box $\mathbf{b}$ and generates LoRA updates and placement tokens for the
    frozen diffusion model.
    \textbf{(b)} Content module: task embeddings are orthogonalized by Gram--Schmidt
    (Eq.~\ref{eq:gs}), while separate heads generate the LoRA updates
    (Eq.~\ref{eq:emit}), and the output-space regularizer preserves updates for earlier
    concepts (Eq.~\ref{eq:anchor}).
    \textbf{(c)} Placement module: the task embedding and bounding box generate
    placement tokens, which image positions attend to through a box mask and
    per-block gate (Eq.~\ref{eq:gsa}).
    Blue elements are trainable or generated, grey ones frozen, hollow ones parameter-free,
    and dashed ones fixed or training-only.
    }
  \label{fig:method}
\end{figure}

The hypernetwork generates both the concept adaptation and the spatial
conditioning. For each concept, a task embedding
$\mathbf{v}_t \in \mathbb{R}^{d_v}$ is provided to a single hypernetwork
$h_\phi$ (Figure~\ref{fig:method}(a)). The hypernetwork has two components:
the \emph{content module}, which generates the low-rank updates used to
personalize the diffusion model (\S\ref{sec:content}), and the
\emph{placement module}, which generates the spatial conditioning for a
user-specified bounding box (\S\ref{sec:localization}).

\subsection{Content module}
\label{sec:content}
\label{sec:hypernet}

\textbf{Generating the update.}
The hypernetwork must generate a concept-specific low-rank update while keeping
the diffusion backbone frozen. For each adapted layer, we use a separate head $\hl$ consisting of two two-layer MLPs. Each head maps the task embedding to the two matrices defining
the corresponding low-rank update:
\begin{equation}
  \hl(\mathbf{v}_t)
  =
  \bigl(\mathbf{A}_{t,\ell},\mathbf{B}_{t,\ell}\bigr),
  \qquad
  \mathbf{W}_{t,\ell}
  =
  \mathbf{W}_{0,\ell}
  +
  \mathbf{A}_{t,\ell}\mathbf{B}_{t,\ell},
  \label{eq:emit}
\end{equation}
where $\mathbf{W}_{0,\ell}\in\mathbb{R}^{d_{\mathrm{out},\ell}\times d_{\mathrm{in},\ell}}$
denotes the frozen pretrained weight of layer $\ell$. We absorb the scaling factor
$\gamma/r$ from Eq.~\ref{eq:lora} into $\mathbf{A}_{t,\ell}$ and denote the
resulting generated update by
$\Delta\mathbf{W}_{\phi,\ell}(\mathbf{v}_t)
=
\mathbf{A}_{t,\ell}\mathbf{B}_{t,\ell}$,
where the subscript $\phi$ indicates the hypernetwork parameters used to generate the update. Because the hypernetwork generates both matrices, the standard LoRA
zero-initialization cannot be applied directly. Instead, we initialize the
output layer producing $\mathbf{B}_{t,\ell}$ to zero, so that the generated update
is zero at the start of training and the pretrained backbone remains unchanged
(Appendix~\ref{app:details}).

The update should capture the visual identity of the concept rather than the rest
of the training prompt, which may describe the surrounding scene. The four adapted
projections act on different parts of the model, and each is applied independently
at every position, which lets us control where the update takes effect.
$\mathbf{W}^K$ and $\mathbf{W}^V$ process text tokens, mapping each token to a key
and a value that determine how it contributes to the image representation.
Updating them at every token would allow the concept adaptation to affect the entire prompt. We therefore restrict these two to the token positions of the concept's
\emph{class word}, which is its identifier in the prompt, following the
masking strategy of TARA~\citep{peng2025tara}, and leave the remaining text tokens
to the frozen projections. In contrast, $\mathbf{W}^Q$ and $\mathbf{W}^O$ operate on image features:
$\mathbf{W}^Q$ produces a query for each image position, while $\mathbf{W}^O$
maps the resulting attention features back into the layer. Their updates can
therefore be applied at every spatial position. Only the class word thus reaches
the generated update, so we use it as the concept identifier and exclude
descriptive attributes from the captions (Appendix~\ref{app:attr}).

\textbf{Preserving earlier concepts.}
Although the layer-specific heads separate the generation of updates across
layers, all concepts still share the same hypernetwork parameters $\phi$.
Updating $\phi$ for a new concept can change the mapping at the inputs
corresponding to previously learned concepts, altering their generated updates.
We reduce this interference in two complementary ways. First, we assign each
concept a distinct direction in the hypernetwork's input space, so mutually orthogonal inputs represent different
concepts. Second, we constrain the
hypernetwork outputs at the inputs of earlier concepts, explicitly preserving
the adaptations learned for them during subsequent training.

We represent each concept with a task embedding in a fixed $d_v$-dimensional
space. For each concept $t$, we sample an initial embedding
$\tilde{\mathbf{v}}_t\sim\mathcal{N}(\mathbf{0},\mathbf{I}_{d_v})$ and
orthogonalize it against the directions assigned to earlier concepts following
the Gram--Schmidt procedure:
\begin{equation}
  \mathbf{v}_t
  =
  \tilde{\mathbf{v}}_t
  -
  \sum_{i<t}
  \langle\tilde{\mathbf{v}}_t,\mathbf{z}_i\rangle\,\mathbf{z}_i,
  \qquad
  \mathbf{z}_t
  =
  \frac{\mathbf{v}_t}{\lVert\mathbf{v}_t\rVert}.
  \label{eq:gs}
\end{equation}
Here $\mathbf{I}_{d_v}$ is the $d_v$-dimensional identity,
$\langle\cdot,\cdot\rangle$ the Euclidean inner product, $\lVert\cdot\rVert$ the
Euclidean norm, and $\mathbf{z}_1,\ldots,\mathbf{z}_{t-1}$ the normalized
embeddings fixed for earlier concepts. Each task consumes one direction, so for
$T\le d_v$ every concept receives a direction orthogonal to all previously
assigned ones. Without the orthogonalization, the updates generated for earlier
concepts drift $7\%$ more over a 49-concept sequence
(Appendix~\ref{app:embeddings}). The second source of interference, arising from
updates to the shared hypernetwork parameters, is addressed by the output-space
regularization described next.

\textbf{Training objective.}
At task $t$, the objective balances learning the new concept with preserving the
updates generated for earlier concepts:
\begin{equation}
  \mathcal{L}_t(\phi)
  =
  \mathcal{L}_{\mathrm{rec}}(\phi)
  +
  \beta\,\mathcal{L}_{\mathrm{output}}(\phi),
  \label{eq:loss}
\end{equation}
where $\beta>0$ controls the trade-off and the second term is present only for
$t>1$. The reconstruction term is the denoising
objective of Eq.~\ref{eq:ldm} evaluated on the current concept using the generated
update:
\begin{equation}
  \mathcal{L}_{\mathrm{rec}}(\phi)
  =
  \mathbb{E}_{\mathbf{x}\sim\mathcal{D}_t,\boldsymbol{\varepsilon},s}
  \left[
    \left\|
    \boldsymbol{\varepsilon}_\theta
    \bigl(\mathbf{x}_s,s,\mathbf{c};\,h_\phi(\mathbf{v}_t)\bigr)
    -
    \boldsymbol{\varepsilon}
    \right\|_2^2
  \right].
  \label{eq:rec}
\end{equation}
Here, $h_\phi(\mathbf{v}_t)$ produces a low-rank update
$\Delta\mathbf{W}_{\phi,\ell}(\mathbf{v}_t)$ for each of the $L$ adapted
cross-attention projections. For each projection $\ell$, its pretrained weight
$\mathbf{W}_{0,\ell}$ is replaced by
$\mathbf{W}_{0,\ell}+\Delta\mathbf{W}_{\phi,\ell}(\mathbf{v}_t)$ when evaluating
the UNet for concept $t$. All other UNet weights remain frozen.

To preserve earlier concepts, we follow the output-space formulation of
\citet{vonoswald2020hypernet} and compare their generated updates with those
produced by a snapshot $\phi_t^\star$ of the hypernetwork taken at the beginning
of task $t$. We then penalize changes in the updates generated for previously
learned concepts,
\begin{equation}
  \mathcal{L}_{\mathrm{output}}(\phi)
  =
  \frac{1}{(t-1)L}
  \sum_{i<t}\sum_{\ell=1}^{L}
  \frac{
    \left\|
      \Delta\mathbf{W}_{\phi,\ell}(\mathbf{v}_i)
      -
      \Delta\mathbf{W}_{\phi_t^\star,\ell}(\mathbf{v}_i)
    \right\|_F^2
  }{
    d_{\mathrm{in},\ell}d_{\mathrm{out},\ell}
  },
  \label{eq:anchor}
\end{equation}
where $\lVert\cdot\rVert_F$ is the Frobenius norm and
$d_{\mathrm{out},\ell}\times d_{\mathrm{in},\ell}$ is the shape of layer $\ell$.
The normalization by the number of parameters makes each layer contribute a
mean squared change per weight entry, so the same $\beta$ applies to
layers of different sizes.

\subsection{Placement module}
\label{sec:localization}
\textbf{Placement tokens.}
The content module determines a concept's visual appearance but not its spatial
location. We therefore derive spatial conditioning from the same task embedding
$\mathbf{v}_t$ and inject it through the frozen backbone's cross-attention layers.
Specifically, we generate \emph{placement tokens} conditioned on the concept and
the desired target region. We specify the target region by a bounding box
$\mathbf{b}=(b_x,b_y,b_w,b_h)$, where $(b_x,b_y)$ denotes its center and
$(b_w,b_h)$ its width and height, normalized to $[0,1]$ by the corresponding
image dimensions. We encode the box using a sinusoidal positional encoding
$\mathrm{PE}(\mathbf{b})$~\citep{vaswani2017attention,tancik2020fourier} and
concatenate it with $\mathbf{v}_t$. The hypernetwork uses the same parameters
$\phi$ to generate both the LoRA updates described in \S\ref{sec:content} and
the placement tokens, with all corresponding heads trained jointly. We denote
the token-generating head by $\htok$, so that
\begin{equation}
  \mathbf{e}_t(\mathbf{b})
  =
  \htok\bigl([\,\mathbf{v}_t\,;\,\mathrm{PE}(\mathbf{b})\,]\bigr)
  \in \mathbb{R}^{M\times d_e},
  \label{eq:gtok}
\end{equation}
where $[\,\cdot\,;\,\cdot\,]$ denotes concatenation, yielding $M$ tokens of
dimensionality $d_e$. Rather than appending these tokens to the prompt, we process
them in a separate attention branch within each cross-attention block, as
described next.

\textbf{Placement attention.}
In each cross-attention block of the pretrained UNet, the image features form
queries through $\mathbf{W}^Q$ and attend to the text tokens $\mathbf{c}$ through
keys and values $\mathbf{W}^K\mathbf{c}$ and $\mathbf{W}^V\mathbf{c}$, followed by
the output projection $\mathbf{W}^O$. We add a parallel attention operation to the
same block whose keys and values are computed from the placement tokens
$\mathbf{e}_t(\mathbf{b})$. Both operations use the same layer input, and their
outputs are summed before $\mathbf{W}^O$, leaving the original text pathway
unchanged. We refer to these two operations as the \emph{text branch} and
\emph{placement branch}.

Since the placement branch operates on generated tokens rather than text
embeddings, it uses its own projections
$\widetilde{\mathbf{W}}^{Q}_{j},\widetilde{\mathbf{W}}^{K}_{j},
\widetilde{\mathbf{W}}^{V}_{j}$ of dimensionality $d_k$ and an output projection
$\widetilde{\mathbf{W}}^{O}_{j}$ mapping back to the width of the text-branch
attention output, so that the two can be summed, for each cross-attention block
$j$. These projections are shared across concepts. To condition the branch on
the concept, the value representation is modulated by a FiLM scale and
shift~\citep{perez2018film},
$(\boldsymbol{\gamma}_F,\boldsymbol{\beta}_F)=\hfilm(\mathbf{v}_t)\in\mathbb{R}^{d_k}\times\mathbb{R}^{d_k}$,
generated from $\mathbf{v}_t$ by a single linear mapping shared across blocks. Let
$p$ denote a position in the UNet latent grid and $\mathbf{x}_p$ its layer input.
The resulting placement attention is
\begin{equation}
  \mathbf{r}_{j,p} = \widetilde{\mathbf{W}}^{O}_{j}
  \Bigl[ (1 + \boldsymbol{\gamma}_F) \odot \widetilde{\mathbf{W}}^{V}_{j} \mathbf{e}_t(\mathbf{b}) + \boldsymbol{\beta}_F \Bigr]
  \,\mathrm{softmax}\!\left( \frac{ \bigl(\widetilde{\mathbf{W}}^{Q}_{j} \mathbf{x}_p\bigr)^{\!\top} \widetilde{\mathbf{W}}^{K}_{j} \mathbf{e}_t(\mathbf{b}) }{ \sqrt{d_k} } \right)^{\!\top} ,
  \label{eq:read}
\end{equation}
where $\odot$ denotes elementwise multiplication and the softmax is taken over the
$M$ placement tokens. This design makes two choices
that are important for spatial control. First, we use attention to read the
placement tokens rather than broadcasting a single vector, allowing different
positions within the box to attend to different tokens. Second, we add the placement branch in parallel with the text branch rather than
applying it before the text attention. Serial injection can suppress the text
pathway~\citep{jang2024reground}, whereas the parallel design preserves the
original text attention and adds the spatial signal to its output.

\textbf{Masking and gating.}
Eq.~\ref{eq:read} produces a placement contribution at every latent position, but
its effect should be restricted to the target region. Let
$(u_p,v_p)\in[0,1]^2$ denote the center of position $p$ in normalized image
coordinates. We therefore weight $\mathbf{r}_{j,p}$ by a soft mask:
\begin{equation}
  \mathrm{inside}(p,\mathbf{b}) =
  \sigma_\lambda\!\left(u_p - b_x + \frac{b_w}{2}\right)
  \sigma_\lambda\!\left(b_x + \frac{b_w}{2} - u_p\right)
  \sigma_\lambda\!\left(v_p - b_y + \frac{b_h}{2}\right)
  \sigma_\lambda\!\left(b_y + \frac{b_h}{2} - v_p\right),
  \label{eq:inside}
\end{equation}
with $\sigma_\lambda(z)=(1+e^{-\lambda z})^{-1}$. The four factors encode the
signed distances to the four box boundaries, yielding a mask that is close to
one inside the box and decays smoothly across its boundary. The sharpness
$\lambda>0$ sets the width of that transition
We then apply the
mask and a learnable per-block gate to the placement contribution:
\begin{equation}
  \mathrm{out}_{j,p}
  =
  \mathrm{out}^{\mathrm{text}}_{j,p}
  +
  \kappa\,\tanh(g_{j})\,
  \mathrm{inside}(p,\mathbf{b})\,\mathbf{r}_{j,p},
  \label{eq:gsa}
\end{equation}
where $\mathrm{out}^{\mathrm{text}}_{j,p}$ is the output of the original text
cross-attention at block $j$ and position $p$, and $\mathrm{out}_{j,p}$ is
the combined output passed to $\mathbf{W}^O$. The scalar gate $g_j$ controls
the strength of the placement branch and is initialized to zero, so the pretrained
text pathway is unchanged at initialization, and $\kappa$ is a scalar gain that we
set to $1$ during training.

\textbf{Training objective.}
Let $\psi$ denote the parameters of the placement branch, comprising the
per-block projections
$\widetilde{\mathbf{W}}^{Q}_{j},\widetilde{\mathbf{W}}^{K}_{j},
\widetilde{\mathbf{W}}^{V}_{j},\widetilde{\mathbf{W}}^{O}_{j}$ and the gates
$g_j$. The heads $\htok$ and $\hfilm$ are part of the hypernetwork and therefore
included in $\phi$. At task $t$, we optimize both sets of parameters jointly:
\begin{equation*}
  \mathcal{L}_t(\phi,\psi)
  =
  \mathcal{L}_{\mathrm{rec}}(\phi,\psi)
  +
  \beta\,\mathcal{L}_{\mathrm{output}}(\phi).
\end{equation*}
This extends Eq.~\ref{eq:loss} by evaluating the reconstruction loss with both
the generated low-rank updates and the placement branch active. The
output-space regularizer in Eq.~\ref{eq:anchor} remains applied only to the
generated low-rank updates and therefore depends on $\phi$ alone. All remaining
implementation details, including the training and sampling procedures, are
provided in Appendix~\ref{app:details}.

\section{Experiments}
\label{sec:results}

\textbf{Setup.}
We use the CIFC benchmark of \citet{dong2024continually}, with ten concepts
(seven objects and three styles) learned sequentially from four to seven
photographs each. Unless stated otherwise, results are averaged over three
training seeds and reported with their standard deviation. Methods are compared
at matched text alignment, and differences below $0.0091$ DINO are treated as
within seed-to-seed variation. We use SD-1.5 unless stated otherwise; results
on SDXL-base-1.0 are reported in Appendix~\ref{app:sdxl}. Further details are
provided in Appendix~\ref{app:details}.

\begin{figure}[t]
  \centering\setlength{\abovecaptionskip}{4pt}
  \includegraphics[width=0.57\linewidth]{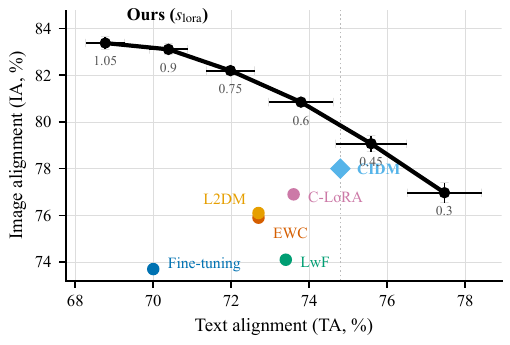}\hfill
  \includegraphics[width=0.40\linewidth]{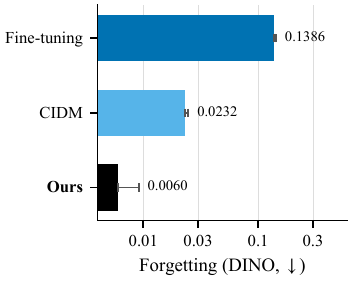}
  \caption{\textbf{Left:} IA versus TA on SD-1.5 for ten
concepts. Baseline results are from the CIFC benchmark~\citep{dong2024continually};
our curve varies the LoRA scale $s_{\mathrm{lora}}$. \textbf{Right:} Forgetting,
measured as the drop in DINO similarity from each concept's peak to its final
score, evaluated at comparable TA. Lower is better.}
  \label{fig:tradeoff}
\end{figure}

\textbf{Generated and independently trained LoRAs.}
Figure~\ref{fig:tradeoff} compares our method with the baselines. On SD-1.5,
our method achieves $79.9 \pm 0.8$ IA at the text alignment of CIDM, compared with $78.0$ for CIDM. The curve in
Figure~\ref{fig:tradeoff} (left) is obtained by varying the LoRA scale
$s_{\mathrm{lora}}$, with the mean and standard deviation computed over three
seeds. The right panel evaluates forgetting at comparable TA: our method at
$s_{\mathrm{lora}}=0.45$ (TA $75.6$), CIDM at $\alpha=0.8$ (TA $75.9$), and
sequential fine-tuning at $s_{\mathrm{lora}}=0.6$ (TA $75.5$). On SDXL, our
method reaches IA $2.4$ points below CIDM after retuning the regularizer weight
(Appendix~\ref{app:sdxl}).

\begin{figure}[t]
  \centering\setlength{\abovecaptionskip}{4pt}
  \input{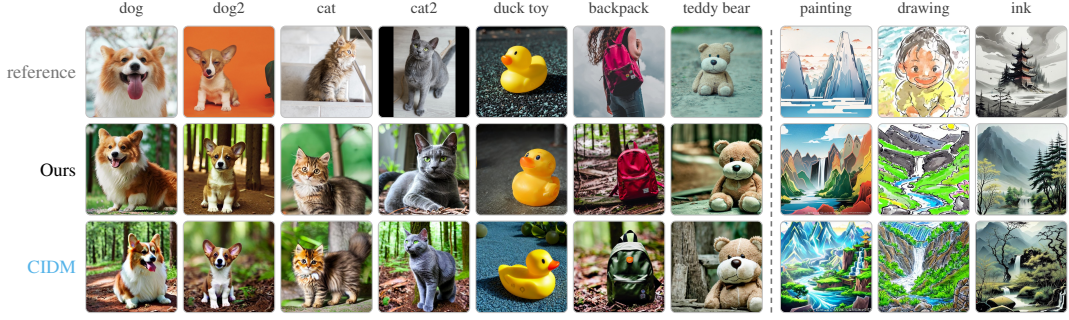}
  \caption{Qualitative comparison on the ten concepts of CIFC. Each column shows
    the reference image followed by generations from our method and CIDM. Both
    methods use SD-1.5, the same sampler and prompts, and the same initial noise,
    so differences between the two rows reflect the respective methods. The first
    seven columns contain object concepts, while the final three contain style
    concepts: painting, drawing, and ink.}
  \label{fig:cidmpair}
\end{figure}

We further compare our generated LoRAs with an oracle in which the ten concepts
are trained independently with dedicated LoRAs. On SD-1.5, our method achieves
$79.1$ IA and $0.617$ DINO, compared with $78.5$ IA and $0.594$ DINO for the
independently trained LoRAs (Appendix~\ref{app:oracle}). This shows that the
generated LoRAs preserve concept identity comparably to independently trained
ones, without requiring a separate LoRA for each concept. Figure~\ref{fig:cidmpair}
provides a qualitative comparison.

\textbf{Forgetting over sequence of tasks.}
We measure forgetting as the drop in DINO similarity from each concept's best
score during the stream to its final score. Our method exhibits a forgetting
of $0.0060 \pm 0.0031$, compared with nearly four times higher forgetting for
CIDM and twenty-three times higher forgetting for sequential fine-tuning
(Figure~\ref{fig:tradeoff}, right). Both CIDM and sequential fine-tuning also
learn a separate token embedding for each concept. These embeddings are
preserved across subsequent tasks, providing an additional persistent
representation of each concept.

\begin{figure}[t]
  \centering\setlength{\abovecaptionskip}{4pt}
  \input{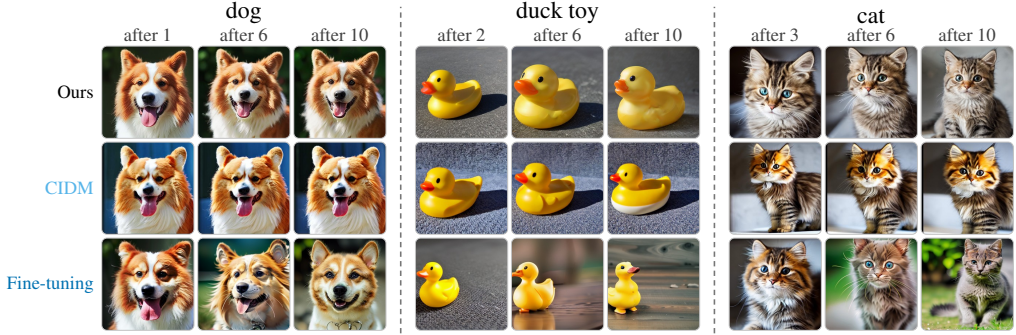}
  \caption{Qualitative comparison of concept        retention during the ten-concept
    sequence. For each of the first three concepts, we show generations immediately
    after its introduction, after task $6$, and after task $10$. All methods use the
    same experimental setting as in Figure~\ref{fig:tradeoff} (right). Our method
    preserves the concept identity throughout the sequence, while sequential
    fine-tuning progressively degrades it. CIDM retains the concepts but can
    introduce features from other concepts through LoRA fusion, as visible in the
    white base of the duck.}
  \label{fig:cmpforget}
\end{figure}

The output-space regularizer accounts for a substantial part of this stability.
Removing it increases forgetting from $0.0094$ to $0.0213$. Even without the regularizer, our method
exhibits six times less forgetting than sequential fine-tuning under the same
backbone and training budget. Figure~\ref{fig:cmpforget}
compares the three methods from the same initial noise, while
Figures~\ref{fig:forgetgrid} and~\ref{fig:forget10} show the full stream.

\begin{figure}[t]
  \centering\setlength{\abovecaptionskip}{4pt}
  \includegraphics{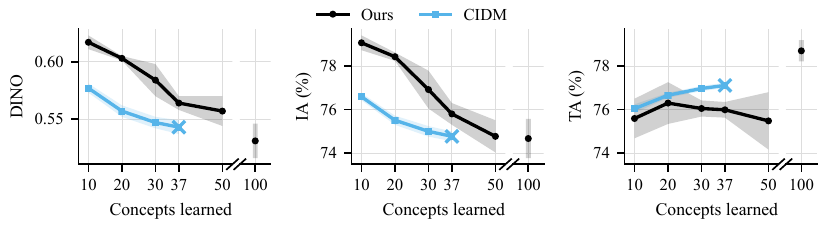}
  \caption{Retention of the first ten concepts over streams of up to one hundred
    concepts. The same ten concepts are generated using models saved after
    $T=10$, $20$, $30$, $37$, $50$, and $100$ tasks; a decrease therefore indicates
    forgetting of the corresponding concept. Results show the mean and standard
    deviation over three training seeds. Our method achieves higher retention for
    seven, seven, five, and four of the ten concepts at $T=10$, $20$, $30$, and
    $37$, respectively. The CIDM results at $T=37$ are taken directly from the
    original work, which reports results up to this stream length. Our method
    continues to $T=100$ without intermediate evaluation between $T=50$ and
    $T=100$, hence the break in the axis.}
  \label{fig:retention}
\end{figure}

\textbf{Performance on longer sequences of concepts.}
We extend the stream to fifty and then one hundred concepts using
CustomConcept101~\citep{kumari2023customdiffusion}. Our hypernetwork has a fixed
size as the stream grows and maintains higher retention at all stream lengths
reported for CIDM (Figure~\ref{fig:retention}). The DINO similarity of the
first ten concepts decreases from $0.617$ after ten tasks to $0.557$ after
fifty, at the same TA, narrowing the gap at longer streams. The reported CIDM
results end at $T=37$, where its CLIP-based adapter weighting reaches the
$77$-token prompt limit. Our method continues to $T=100$, with the first ten
concepts retaining $0.531$ DINO similarity.

The parameter cost also remains constant: our hypernetwork has the same size at
every stream length, whereas storing one LoRA per concept increases the
parameter count with the number of concepts (Figure~\ref{fig:scaling}). The
generated LoRAs add $14\%$ to the frozen backbone, compared with $45\%$ for
CIDM at $T=10$ and $154\%$ at $T=37$.

\begin{figure}[t]
  \centering\setlength{\abovecaptionskip}{4pt}
  \input{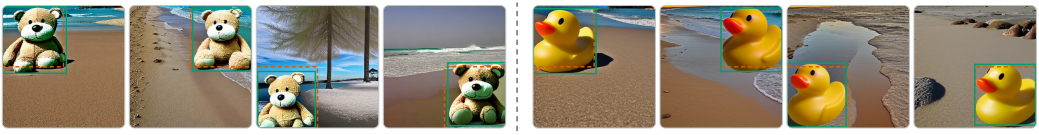}
  \caption{Spatial localization for two concepts, a teddy bear (left) and a duck
    (right), each generated at four requested image locations. The dashed orange
    box marks the target bounding box provided to the model, while the solid green
    box shows the bounding box detected by Mask R-CNN on the generated image. The
    same beach scene and lighting are maintained across the generated images, while
    only the target concept and its requested location vary. The complete
    evaluation grid is shown in Figure~\ref{fig:grids}.}
  \label{fig:placehalf}
\end{figure}

\textbf{Spatial localization from few-shot concepts.}
We evaluate placement with an object detector, measuring both whether the concept
appears in the requested quadrant and whether its detected box overlaps the
target box (Appendix~\ref{sec:pitfalls}). Without the placement module, the box
is ignored: specifying the location in the prompt does not improve placement,
while a training-free regional layout loses the subject in one quarter of
generations. With our module, the concept appears in the requested quadrant in
every generation, and $96\%$ of detected generations achieve an IoU above
$0.5$. The spatial constraint also improves identity: DINO similarity on the
subject mask increases from $0.599$ without a box to $0.717$ with one
(Table~\ref{tab:grounding}). Figure~\ref{fig:placehalf} shows two concepts in
each quadrant, while Figure~\ref{fig:grids} shows the complete grid.

\section{Conclusions}
\label{sec:conclusion}

We presented a continual personalization framework that uses a single fixed-size
hypernetwork to generate concept-specific adaptations for a frozen diffusion
model. Each concept is represented by a compact task embedding, while
output-space regularization preserves the adaptations generated for previously
learned concepts as the hypernetwork is updated. The framework also supports
spatial control, allowing the learned concept to be placed within a
user-specified bounding box. Across the benchmark, our approach achieves
competitive personalization quality while substantially reducing forgetting and
keeping the model-side storage constant as new concepts are acquired. These
results show that a single hypernetwork can generate the required adaptation for
each concept, avoiding the need to store separate parameters for every concept.

\textbf{Limitations.}
The method has three practical limitations. First, image quality depends on the
number of reference images available for each concept, with fewer examples
providing less information for personalization. Second, the hypernetwork has
finite representational capacity, which may become a limiting factor as the
number and diversity of concepts in the task sequence grow substantially.
Finally, performance depends on the underlying diffusion backbone, and the
hypernetwork and regularization settings may require adjustment when transferred
to a different architecture.

\section*{Ethics statement}

The two benchmarks we use contain photographs of people. CustomConcept101 \citep{kumari2023customdiffusion} includes person concepts, and several object concepts in the benchmark of \citet{dong2024continually} are photographed being worn or held. We use both datasets only as released, learn no concept outside them, and release no images of people with the code.

Personalization of a text-to-image model is dual use: the same mechanism that lets a photographer re-render their own subject lets someone synthesize a likeness without consent. Our contribution is to learning many concepts in sequence in one network rather than to the fidelity of any one of them, so it does not make a likeness easier to forge than the per-concept adapters it replaces. It does make holding many of them at once cheaper, and a store of a hundred identities in one fixed-size network is easier to distribute than a hundred separate adapters. The placement module of \S\ref{sec:localization} carries the same dual use at the level of composition: placing a subject in a chosen part of a scene is what a photographer wants and also what a fabricated photograph needs. We report both capabilities, including where they fail (Appendix~\ref{sec:transfer}), because a reader weighing that risk needs the limits as much as the results.

\section*{Reproducibility statement}
The code,
configurations for all reported runs, evaluation scripts, and placement test are
available in a public repository at
\url{https://github.com/genwro-ai/clasp}. We also release the
hundred-concept sequence of \S\ref{sec:results}, including the
concepts, their order, captions, and prompts. The images are drawn from the
original datasets and are not redistributed. The experimental protocol and
seeds are described in \S\ref{sec:results}, while Appendix~\ref{app:details}
provides the remaining implementation details, including the optimizer,
schedule, adapter shapes, data augmentation, and compute. Appendix~\ref{app:anchor}
gives the closed form used to evaluate the output-space regularizer.

\section*{Use of large language models}

An LLM assistant was used, under the direction of the authors, to draft and
revise the prose of this paper and to assist with the code used to produce its
figures. The training captions for the ninety additional concepts were
generated by an LLM from the corresponding images, following the protocol and
verification procedure described in Appendix~\ref{app:data}. The research questions, method,
experiments, analysis, and scientific claims are the authors' responsibility.

\bibliography{iclr2027_conference}
\bibliographystyle{iclr2027_conference}

\newpage
\appendix
{\Large{Appendix}}
\vspace{1em}

Appendix~\ref{app:related} extends the related work, and Appendix~\ref{app:ldm} recalls latent diffusion and classifier-free guidance. Appendix~\ref{app:details} gives the training and generation settings, and Appendix~\ref{app:anchor} the efficient computation of the output-space regularizer. Appendix~\ref{app:data} describes the captions of the added concepts and why attributes are stripped from them, and Appendix~\ref{app:metrics} defines the metrics and what they cannot resolve. For the content module, Appendices~\ref{app:curve}, \ref{app:oracle} and~\ref{app:sdxl} give the numbers behind the trade-off curve, the comparison with independently trained adapters and the second backbone. Appendix~\ref{app:ablations} collects the ablations, Appendix~\ref{app:forgetstream} shows the concepts along the sequence, and Appendix~\ref{app:embeddings} analyzes the task embeddings. The placement module is evaluated in Appendix~\ref{app:grids}, and Appendix~\ref{sec:composition} shows the composition of several concepts.

\section{Extended related work}
\label{app:related}

\textbf{Continual learning and personalization.}
Continual learning studies models trained on a sequence of tasks while retaining previously acquired knowledge, with catastrophic forgetting as its central challenge~\citep{mccloskey1989catastrophic,de2021continual,wang2024continualsurvey}. Existing approaches mitigate forgetting through \emph{regularization}, which constrains changes to parameters or functions important for previous tasks~\citep{kirkpatrick2017ewc,li2018lwf,Aljundi2017MemoryAS,zenke2017si}, \emph{rehearsal}, which stores or generates data from previous tasks~\citep{rebuffi2017icarl,castro2018end,hou2019learning,douillard2020podnet,shin2017continual,wu2018memory,cong2020gan,van2020brain,yin2020dreaming,smith2021always,gao2022r,qiu2024dcmi,tongmodel}, and \emph{parameter-efficient adaptation}, which learns task-specific prompts, adapters, or low-rank updates while keeping a pretrained backbone fixed~\citep{Wang2021LearningTP,wang2022dualprompt,smith2023codap,yu2024moeadapter,liang2024inflora}. Most of this literature focuses on discriminative models, where forgetting is measured by degradation in accuracy on previously learned classes. Our setting instead considers generative models, where each task corresponds to a visual concept, and forgetting manifests as a degradation in the ability to reproduce previously learned concepts under novel prompts. Diffusion personalization adapts a pretrained text-to-image model to a specific subject or style from a small set of reference images using learned token embeddings~\citep{gal2023textualinversion}, full or partial fine-tuning~\citep{ruiz2023dreambooth,kumari2023customdiffusion}, or low-rank adapters~\citep{hu2022lora}. Continual personalization extends this setting to a sequence of concepts, introducing the additional challenge of preserving previously learned concepts as new ones are acquired. Existing approaches address this through concept-specific isolation~\citep{smith2023continual,smith2024continual}, replay or distillation~\citep{sun2024create,jha2025mining}, and parameter merging or allocation~\citep{staniszewski2024low,liao2025continual}. CIDM~\citep{dong2024continually,dong2026crafting}, whose CIFC benchmark we adopt, maintains a separate set of low-rank parameters for each concept, causing its parameter storage to grow with the sequence length. CNS~\citep{liao2025continual} avoids this growing storage by allocating subsets of the backbone to individual concepts, but consequently limits the number of concepts by the available parameter capacity. Our approach instead keeps the diffusion backbone fixed and uses a single \emph{hypernetwork} to generate concept-specific adaptations for an increasing number of concepts.

\textbf{Hypernetworks and concept separation.}
Hypernetworks~\citep{Ha2017HyperNetworks} generate the parameters of a target network conditioned on a compact task representation. In continual learning, \citet{vonoswald2020hypernet} show that regularizing changes in the generated parameters, rather than the hypernetwork weights themselves, can substantially reduce forgetting. We adopt this output-space perspective for continual diffusion personalization. Hypernetworks have also been explored for fast diffusion personalization~\citep{ruiz2023hyperdreambooth} and concept removal~\citep{wojcik2026unhype}, but these methods do not consider sequential acquisition of concepts or the interference that arises when a shared hypernetwork is repeatedly updated. A complementary line of work addresses concept interference by separating concept-specific adaptations in weight space. Orthogonal Adaptation~\citep{po2024orthogonal} and SeqLoRA~\citep{parsa2026seqlora} constrain adapters of different concepts to occupy distinct subspaces, while related interference-free low-rank adaptation has been studied in discriminative continual learning~\citep{liang2024inflora}. These approaches separate concepts directly in parameter space, so the capacity for storing distinct concepts remains tied to the available adaptation parameters. In contrast, our method uses a single hypernetwork shared across all concepts. Each concept is represented by a distinct task embedding, allowing the hypernetwork to generate a different adaptation for each concept while keeping its own size fixed. TARA~\citep{peng2025tara} addresses a different source of interference by restricting a concept-specific update to the token positions associated with that concept, rather than allowing it to affect the entire prompt.

\section{Background: latent diffusion models}
\label{app:ldm}

A latent diffusion model~\citep{rombach2022ldm} generates images by denoising a
sample from a Gaussian prior in the latent space of a pretrained autoencoder.
A frozen text encoder maps a natural-language prompt to a sequence of text-token
embeddings $\mathbf{c} \in \mathbb{R}^{N\times d_c}$, where $N$ is the number of
tokens and $d_c$ is the embedding dimension. Given an image with latent
representation $\mathbf{x}$, the forward diffusion process produces $\mathbf{x}_s
=
\sqrt{\bar\alpha_s}\,\mathbf{x}
+
\sqrt{1-\bar\alpha_s}\,\boldsymbol{\varepsilon},$
where $\boldsymbol{\varepsilon} \sim \mathcal{N}(\mathbf{0},\mathbf{I})$ is
standard Gaussian noise, $\mathbf{I}$ is the identity matrix,
$\{\bar\alpha_s\}_{s=1}^{S}$ is a fixed noise schedule, and $S$ denotes the
total number of diffusion timesteps. The timestep $s$ is sampled uniformly
from $\{1,\ldots,S\}$. We reserve $t$ for the concept index. The denoising
network is trained to recover the noise added at each timestep. Specifically,
a UNet~\citep{ronneberger2015unet} $\boldsymbol{\varepsilon}_\theta$ with
parameters $\theta$ predicts $\boldsymbol{\varepsilon}$ from the noisy latent,
timestep, and text conditioning by minimizing
\begin{equation}\label{eq:ldm}
\mathcal{L}_{\mathrm{LDM}}
=
\mathbb{E}_{\mathbf{x},\boldsymbol{\varepsilon},s}
\left[
\left\|
\boldsymbol{\varepsilon}_\theta(\mathbf{x}_s,s,\mathbf{c})
-
\boldsymbol{\varepsilon}
\right\|_2^2
\right].
\end{equation}
At inference, classifier-free guidance~\citep{ho2021classifierfree} combines
conditional and unconditional predictions as
\begin{equation}
\tilde{\boldsymbol{\varepsilon}}
=
\boldsymbol{\varepsilon}_\theta(\mathbf{x}_s,s,\varnothing)
+
w\left(
\boldsymbol{\varepsilon}_\theta(\mathbf{x}_s,s,\mathbf{c})
-
\boldsymbol{\varepsilon}_\theta(\mathbf{x}_s,s,\varnothing)
\right),
\label{eq:cfg}
\end{equation}
where $\tilde{\boldsymbol{\varepsilon}}$ is the guided noise prediction, $w$ is
the guidance scale, and $\varnothing$ denotes the null text condition obtained
by encoding an empty prompt. The text embeddings $\mathbf{c}$ condition
the UNet through cross-attention, whose query, key, value, and output
projections are denoted by $\mathbf{W}^Q$, $\mathbf{W}^K$, $\mathbf{W}^V$, and
$\mathbf{W}^O$, respectively. Personalization methods adapt some or all of
these projections, and we adapt all four.

\section{Training and generation details}
\label{app:details}

\paragraph{Backbones and parameters.} We use two backbones, SD-1.5 at $512^2$ and SDXL-base-1.0 at
$1024^2$. The UNet, the VAE and the text encoders are frozen, and only the hypernetwork trains. It
has $26.0$M parameters on SD-1.5 and $191.1$M on SDXL. The content heads account for $21.9$M and
$167.5$M of them, over $L = 64$ and $L = 280$ adapted layers, and the placement module for the
remaining $4.1$M and $23.5$M. SD-1.5 conditions its cross-attention on the last hidden state of a
CLIP ViT-L/14 text encoder~\citep{radford2021clip}, of width $768$, which is also the width of the
placement tokens. SDXL-base-1.0 conditions on the concatenated penultimate hidden states of its two
text encoders and on the pooled projection of the second, of width $1280$.

\paragraph{Training.} Each task is trained for $800$ steps with batch size $2$, AdamW at learning
rate $10^{-4}$ and weight decay $10^{-2}$, and the lookahead rate $\eta = 10^{-4}$. Following \citet{vonoswald2020hypernet}, the output-space
regularizer of Eq.~\ref{eq:anchor} is evaluated at the lookahead parameters
$\phi - \eta\nabla_\phi\mathcal{L}_{\mathrm{rec}}$ rather than at $\phi$. The regularizer weight is $\beta = 7120$ on SD-1.5 and $\beta = 3000$ on SDXL.
It is the only hyperparameter that does not transfer between the two backbones, and
Appendix~\ref{app:sdxl} shows what carrying it over unchanged costs. The hypernetwork trains in
fp32. The frozen backbone is kept in fp32 on SD-1.5 and in bf16 on SDXL, and the VAE in fp32 on
both.

\paragraph{Content module.} We adapt the $\mathrm{attn2}.\{\mathrm{to\_q}, \mathrm{to\_k},
\mathrm{to\_v}, \mathrm{to\_out}.0\}$ projections of every attention block and generate adapters at
rank $r = 4$. Each head consists of two two-layer MLPs of hidden width $50$, one per factor, and each
layer has a learned scalar gain folded into $\mathbf{A}$. Every linear layer of a head draws its
weights and biases uniformly from $[-1/\sqrt{n_{\mathrm{in}}}, 1/\sqrt{n_{\mathrm{in}}}]$ (the
default of a PyTorch linear layer), with $n_{\mathrm{in}}$ its input width ($256$ for the first
layer, $50$ for the second). The exception is the output layer that produces $\mathbf{B}$, whose
weights and bias are set to zero, and the gain starts at one. The generated update is therefore
exactly zero while $\mathbf{A}$ is not, so the first gradient reaches the $\mathbf{B}$ branch, as for
a LoRA whose $\mathbf{B}$ starts at zero. The network is initialized once, before the first concept,
and every later concept starts from the parameters its predecessors left. The regularizer adds two
hypernetwork passes over the old embeddings per training step.

Task embeddings are $d_v = 256$ wide. The initial task embedding $\tilde{\mathbf{v}}_t$ of concept $t$ is drawn from seed $1234 + t - 1$, the
same in every training seed, so the basis of Eq.~\ref{eq:gs} can be rebuilt from the task order
alone. The checkpoint stores two vectors per concept, $0.03$M parameters at $T = 50$, but neither
carries information. The initial task embedding follows from the index, and the orthogonalized one is what
replaying Gram--Schmidt returns. What
a concept adds to the store is therefore an integer.

\paragraph{Placement module.} The box mask of Eq.~\ref{eq:inside} uses $\lambda = 40$, which makes its
edge about a tenth of the frame wide. The positional encoding $\mathrm{PE}(\mathbf{b}) \in \mathbb{R}^{64}$ is
the sine and cosine of each of the four box coordinates at eight frequencies. The token head
generates $M = 4$ tokens of the width of the text embeddings, and the placement branch projects
queries, keys and values to $d_k = 64$ dimensions.

\paragraph{Training composites.} The placement branch cannot be trained on the concept's photographs
alone. In a photograph the subject is wherever it happens to be, and the noisy latent already
reveals that, so the box adds no information. On half of the training steps we therefore replace
the photograph with a composite, and only in the high-noise half of the schedule,
$s \in [S/2, S)$, where layout is decided and the appearance of the composite is not yet resolved. To build a composite, we segment the subject, shrink its alpha matte
by three pixels, and paste it onto one of about one hundred backgrounds synthesized once by the
frozen backbone, where it covers $45\%$ to $85\%$ of the frame's long side. Without the shrink, the
soft edge carries the source background into the composite, and at inference it appears as a rim
around every subject. The benchmark's seven object concepts are segmented with
ISNet~\citep{qin2022isnet}, and the added concepts with Grounding-DINO boxes for the class word
refined by SAM~\citep{liu2024groundingdino,kirillov2023sam}. For the $45$ of $515$ photographs where
the automatic mask failed, we chose among the candidate masks by hand and recorded the choice. The
remaining training steps use the concept's own photographs, with a random square crop of $80$ to
$100\%$ of the short side and a random horizontal flip.

\paragraph{Generation.} Generation uses $50$ denoising steps at guidance $7.5$, and for evaluation we
draw ten images per prompt. Bootstrapping, in the manner of
MultiDiffusion~\citep{bartal2023multidiffusion}, replaces everything outside the box by a background
latent for the first $K$ steps, so the subject can form only inside it. With a box, unless
stated otherwise, the gain is $\kappa = 2$, bootstrapping covers the first $K = 15$ steps, and
$s_{\mathrm{lora}} = 0.7$. The generated adapters and the attention branch enter only the conditional
pass of classifier-free guidance, and the unconditional pass runs the frozen backbone, so guidance
amplifies both like any other condition. The gain schedule and the bootstrap each span the first
$30\%$ of the denoising steps. The background latent is generated in ten steps with both modules
off, so it agrees with the scene and needs no pre-rendered images. The box is dilated by three
latent cells before the replacement. The replacement mask of the bootstrap is binary, because
blending two noise draws leaves the input under-noised for its timestep and desaturates the output.

\paragraph{Compute cost.} Training one ten-concept sequence takes about an hour on a single NVIDIA
GH200 for SD-1.5 and $5.5$ to $7$ hours for SDXL. On the same hardware, CIDM takes $1$ hour $8$
minutes to $1$ hour $53$ minutes for the same ten concepts, and its time per task grows along the
sequence, from $7$ minutes for the first task to $16$ for the ninth. For generation, measured on one
A100 with batch size $5$, $50$ DPM-Solver steps and fp32, CIDM adds $45\%$ to the frozen backbone at
$T = 10$ and $154\%$ at $T = 37$, because every step runs all $T$ adapters. Our method adds $14\%$ at
$T = 10$, $37$ and $100$ alike, including one hypernetwork pass of $14$ ms per generation.
Generation times are medians over $12$ batches. An evaluation sweep takes between $5$ minutes and
about $4$ GPU-hours. All experiments ran on GH200 and A100 nodes. A replication of the inference
protocol on a different cluster agreed with the original to within three of $84$ samples on the
placement metrics and to within $0.0015$ DINO per concept.

\section{Efficient computation of the output-space regularizer}
\label{app:anchor}

Each update $\Delta\mathbf{W}_{\phi,\ell}(\mathbf{v})$ is a $d_{\mathrm{out}} \times d_{\mathrm{in}}$ matrix, up to
$1280 \times 1280$ on SD-1.5, and Eq.~\ref{eq:anchor} compares two of them for every earlier concept,
at every layer and every step. Both are rank-$r$ products, $\Delta\mathbf{W}_1 = \mathbf{A}_1 \mathbf{B}_1$ and
$\Delta\mathbf{W}_2 = \mathbf{A}_2 \mathbf{B}_2$, with the gains folded into $\mathbf{A}$. Their squared
Frobenius distance therefore reduces to traces of $r \times r$ matrices,
\begin{equation}
  \lVert \mathbf{A}_1 \mathbf{B}_1 - \mathbf{A}_2 \mathbf{B}_2 \rVert_F^2
  = \operatorname{tr}(G_{11} H_{11}) - 2\,\operatorname{tr}(G_{12} H_{12}) + \operatorname{tr}(G_{22} H_{22}),
  \label{eq:anchor-trace}
\end{equation}
where $G_{ij} = \mathbf{A}_i^{\!\top} \mathbf{A}_j$ and $H_{ij} = \mathbf{B}_j \mathbf{B}_i^{\!\top}$.
The identity follows from $\lVert X \rVert_F^2 = \operatorname{tr}(X^{\!\top} X)$ and the cyclic
property of the trace. Each Gram matrix costs $O(r^2 d)$ to form, so the regularizer costs
$O\bigl(r^2 (d_{\mathrm{in}} + d_{\mathrm{out}})\bigr)$ per layer instead of
$O(r\, d_{\mathrm{in}} d_{\mathrm{out}})$, and no update is ever materialized. The division by
$d_{\mathrm{in}} d_{\mathrm{out}}$ in Eq.~\ref{eq:anchor} puts the term on the scale of a per-entry
mean squared error, so one value of $\beta$ serves layers of different widths.

\section{Data and captions}
\label{app:data}

\subsection{Caption attributes}
\label{app:attr}

The training captions describe the scene around a concept but not the concept's own attributes,
such as its color or material. An attribute named in the caption competes with the adapter for
carrying it. If every caption of the backpack says ``red'', the text branch already accounts for
the color, and the adapter has no reason to encode it. At inference the prompt is ``a photo of
backpack'' and names no color. The backpack then comes out gray. Stripping the attribute from the
captions moves the color into the adapter.

Table~\ref{tab:attr} tests this on the three concepts whose captions originally named an
attribute, against a variant trained without the stripping. Both are evaluated with the bare
prompt. Stripping raises identity on the duck toy and the backpack, whose captions all named the
attribute. It leaves the cat unchanged, since only one of its five captions named it, and it leaves
the seven concepts without an attribute unchanged. Across all ten concepts, DINO rises from $0.601$
to $0.620$. The effect appears where the mechanism predicts it and nowhere else.

\begin{table}[h]
  \centering
  \caption{Effect of stripping caption attributes. $\Delta$DINO is stripped minus unstripped, final
  checkpoint at $s_\mathrm{lora} = 0.45$, scored with the bare prompt. The unstripped variant is a
  control trained with the same recipe, seed and concepts, differing in the stripping and in being
  configured for ten concepts rather than fifty.}
  \label{tab:attr}
  \begin{tabular}{lc}
    \toprule
    Concept (attribute in caption) & $\Delta$DINO \\
    \midrule
    backpack (``red'', $6/6$ captions)   & $+0.054$ \\
    duck toy (``yellow rubber'', $4/4$)   & $+0.124$ \\
    cat (``fluffy'', $1/5$)               & $-0.002$ \\
    \midrule
    seven concepts without an attribute (mean) & $+0.002$ \\
    \bottomrule
  \end{tabular}
\end{table}

\subsection{Captions for the added concepts}

The benchmark provides one caption per image, and CustomConcept101 provides none. Training on the
two together would confound the length of the sequence with a change of captioning regime.
The captions for the ninety added concepts were therefore written from the images themselves,
following the argument of Appendix~\ref{app:attr}. For each concept, all of its images were
considered together. What stays constant across them (color, material, shape, markings, a face) was
separated from what varies (where the object is, what it rests on, what is next to it), and only
the latter was described.

Each of the $515$ captions contains its class word exactly once, in its base form. The surroundings
may be described, so $32$ captions name an object that is also another concept of the sequence, most
often a table. In three captions a content word directly precedes the class word, and in each case
it describes a posture rather than an attribute (\emph{a seated person}, \emph{a stretching cat},
\emph{a centered toy figure}). For evaluation we use the CIFC prompts and assign each
added concept to one of its four categories.

\section{Metrics}
\label{app:metrics}

We report three metrics. Text alignment (TA, CLIP text-image similarity) and image alignment (IA,
CLIP image-image similarity) are the two the benchmark reports, and we use them to compare with
published work. DINO ViT-S/16 similarity~\citep{caron2021dino} separates identities more finely,
and we use it among our own runs, where no published numbers exist: for forgetting, ablations and
placement. All three follow the CIFC evaluation script: CLIP ViT-B/32 with the CLIPScore
scaling for TA, the mean cosine over all generated-reference pairs for IA, and the same DINO
extractor.

Forgetting is computed from the full $55$-cell matrix of scores of the model after task $k$ on
concept $j \le k$. It is the mean, over the concepts learned before the last task, of
$\max_{j \le k < \mathrm{final}} \mathrm{score}(k, j) - \mathrm{score}(\mathrm{final}, j)$, so a
concept that scores best at the end contributes a negative term.

Identity and TA trade off along $s_{\mathrm{lora}}$, so we compare methods at matched
TA. We claim no difference below $0.0091$ DINO, the largest seed-to-seed standard
deviation we measured across configurations at ten concepts. After fifty tasks the spread grows to
$0.015$ (Table~\ref{tab:curve}).

\subsection{What the metrics cannot resolve}
\label{sec:pitfalls}

\paragraph{Reference-set ceiling.} The references of one concept
already differ from one another. Their mean pairwise DINO similarity ranges from $0.42$ (painting)
to $0.91$ (dog2), and from $0.60$ to $0.91$ among the objects, where the backpack is lowest because
one reference shows it worn by a person. This ceiling is a property of the data, and it orders the
concepts almost exactly as our results do: the correlation between a concept's ceiling and our
DINO on it is $0.93$ (Spearman) and $0.97$ (Pearson), and dog2 is the only concept out of rank. Because a per-concept score read without its ceiling says more about the concept than about the method, Table~\ref{tab:ceiling} lists each score next to its ceiling. At
$s_{\mathrm{lora}} = 0.45$ our model reaches $0.82$ to $0.98$ of the ceiling on the objects and
$0.75$ to $0.80$ on the styles. The remaining gap lies along the trade-off curve. At
$s_{\mathrm{lora}} = 0.9$, the identity end of the same curve, the same checkpoint reaches $0.90$ to
$1.01$ of the ceiling on every concept. At that scale, what separates our scores from the ceiling
is TA traded away on purpose, not identity the adapter cannot reach.

\begin{table}[h]
  \centering
  \caption{Reference-set ceiling per concept, the mean pairwise DINO similarity among the
  concept's own reference photographs, beside our DINO at both ends of the trade-off curve, model
  after all ten tasks. The ceiling varies by more than a factor of two across concepts and orders
  them almost exactly as our results do. At $s_{\mathrm{lora}} = 0.45$ a gap to the ceiling
  remains, and at $s_{\mathrm{lora}} = 0.9$ it closes. Styles sit lowest because a style has no
  single subject for DINO to match.}
  \label{tab:ceiling}
  \footnotesize
  \begin{tabular}{lccccc}
    \toprule
    & & & \multicolumn{2}{c}{ours (ratio to ceiling)} \\
    \cmidrule(lr){4-5}
    Concept & references & ceiling & $s_{\mathrm{lora}} = 0.45$ & $s_{\mathrm{lora}} = 0.9$ \\
    \midrule
    painting      & $7$ & $0.423$ & $0.319$ ($0.75$) & $0.409$ ($0.97$) \\
    ink painting  & $5$ & $0.432$ & $0.347$ ($0.80$) & $0.398$ ($0.92$) \\
    drawing       & $6$ & $0.555$ & $0.421$ ($0.76$) & $0.513$ ($0.92$) \\
    \midrule
    backpack      & $6$ & $0.604$ & $0.546$ ($0.90$) & $0.594$ ($0.98$) \\
    duck toy      & $4$ & $0.721$ & $0.679$ ($0.94$) & $0.670$ ($0.93$) \\
    teddy bear    & $7$ & $0.774$ & $0.745$ ($0.96$) & $0.771$ ($1.00$) \\
    cat2          & $5$ & $0.782$ & $0.768$ ($0.98$) & $0.794$ ($1.01$) \\
    dog           & $5$ & $0.849$ & $0.800$ ($0.94$) & $0.803$ ($0.95$) \\
    cat           & $5$ & $0.898$ & $0.836$ ($0.93$) & $0.869$ ($0.97$) \\
    dog2          & $5$ & $0.913$ & $0.747$ ($0.82$) & $0.825$ ($0.90$) \\
    \bottomrule
  \end{tabular}
\end{table}

\paragraph{Limits of the placement metrics.} Containment also rises when protruding
parts of the subject are removed, and a duplicated subject still yields a well-formed detection
box. We found subject amputation, canvas clipping and cut-out halos only by inspecting image grids,
and none of them showed in the tables. The placement numbers must therefore be read together with
the grids (Figure~\ref{fig:grids}).

\section{Full adapter-scale results}
\label{app:curve}

Table~\ref{tab:curve} lists the values behind the curve in Figure~\ref{fig:tradeoff}, after ten tasks
and after fifty.

\begin{table}[h]
  \centering
  \caption{Our model across LoRA scales, box-free protocol, mean and standard deviation over three training seeds, after ten tasks and after the same model has learned fifty. TA and IA in percent, DINO whole-image. The benchmark reports IA and TA only.}
  \label{tab:curve}
  \footnotesize
  \begin{tabular}{lcccccc}
    \toprule
    & \multicolumn{3}{c}{after task $10$} & \multicolumn{3}{c}{after task $50$} \\
    $s_{\mathrm{lora}}$ & TA $\uparrow$ & IA $\uparrow$ & DINO $\uparrow$ & TA $\uparrow$ & IA $\uparrow$ & DINO $\uparrow$ \\
    \midrule
    $0.30$ & $77.47_{\pm 0.96}$ & $76.97_{\pm 0.42}$ & $0.573_{\pm 0.008}$ & $76.65_{\pm 1.00}$ & $72.94_{\pm 0.66}$ & $0.519_{\pm 0.011}$ \\
    $0.45$ & $75.59_{\pm 0.91}$ & $79.07_{\pm 0.34}$ & $0.617_{\pm 0.006}$ & $75.48_{\pm 1.32}$ & $74.77_{\pm 0.74}$ & $0.557_{\pm 0.013}$ \\
    $0.60$ & $73.79_{\pm 0.83}$ & $80.85_{\pm 0.11}$ & $0.644_{\pm 0.006}$ & $74.10_{\pm 1.54}$ & $76.22_{\pm 0.85}$ & $0.579_{\pm 0.014}$ \\
    $0.75$ & $71.98_{\pm 0.62}$ & $82.20_{\pm 0.05}$ & $0.659_{\pm 0.006}$ & $72.65_{\pm 1.59}$ & $77.25_{\pm 0.84}$ & $0.592_{\pm 0.015}$ \\
    $0.90$ & $70.39_{\pm 0.49}$ & $83.11_{\pm 0.24}$ & $0.663_{\pm 0.006}$ & $71.22_{\pm 1.53}$ & $77.78_{\pm 0.71}$ & $0.592_{\pm 0.015}$ \\
    $1.05$ & $68.77_{\pm 0.50}$ & $83.38_{\pm 0.27}$ & $0.655_{\pm 0.005}$ & $69.63_{\pm 1.52}$ & $77.51_{\pm 0.44}$ & $0.572_{\pm 0.011}$ \\
    \bottomrule
  \end{tabular}
\end{table}

Two things follow from the table. First, identity saturates at the high end of the scale. From
$s_{\mathrm{lora}} = 0.9$ to $1.05$, IA gains only $0.27$, against $0.91$ for the step
before, and DINO falls from $0.663$ to $0.655$. Raising the scale further adds no identity.
Second, after fifty tasks the TA at a given scale moves by less than a point, while
IA falls by four to six points and DINO by $0.05$ to $0.08$. The model therefore loses identity at an unchanged TA.

\section{Comparison with independently trained adapters}
\label{app:oracle}

It is not obvious that a generated adapter can be as good as a trained one. To test this, we train
ten independent rank-$4$ adapters, each directly on the frozen backbone, with the same data and the
same text side as our recipe: no identifier token, the delta applied only at the class-word tokens,
the same captions and augmentation, and $800$ steps per concept. At inference, each adapter is
selected by its task index. This control stores a separate adapter for every concept ($0.40$M parameters each). Its storage grows with $T$, but it cannot forget. At a common TA of
$75.6$, ours at $s_{\mathrm{lora}} = 0.45$, the control reaches an IA of $78.5$ and a
DINO of $0.594$, against $79.1 \pm 0.3$ and $0.617$ for the generated adapters (two seeds for the
control, three for ours). Doubling its training to $1600$ steps per concept leaves it at $78.4$ and
$0.593$. The generated adapters therefore match independently trained ones on this benchmark, from
a network whose size does not depend on $T$.

\section{Results on SDXL}
\label{app:sdxl}

\paragraph{Storage on SDXL.} Since we trained CIDM on SD-1.5 only, we estimate its SDXL store from the
structure of its SD-1.5 models. These adapt exactly the four cross-attention projections we adapt,
at rank $4$, with nothing on self-attention, and each of the fifty concepts carries two tokens whose
embedding is kept once per layer. The same structure on SDXL, with $70$ cross-attention layers and a
$2048$-dimensional context, gives $3.14$M in low-rank factors and $0.29$M in embeddings, or $3.43$M
per concept, against $191.1$M for our network. The two are equal at $T = 56$, earlier than the $62$
on SD-1.5. Applied back to SD-1.5, the same procedure returns $397{,}824$ and $24{,}576$, exactly the
two numbers its SD-1.5 models contain, which supports the estimate. If CIDM kept a separate
embedding for each of SDXL's two text encoders, its store would be larger and the crossing earlier
still, so the assumption favors CIDM.

\paragraph{Transferring the recipe.} On SDXL-base-1.0 the branch and the training recipe transfer
without a change of shape, giving $191.1$M parameters, $167.5$M of them across $L = 280$ adapted
layers. One hyperparameter does not transfer. Carried over unchanged from SD-1.5, the regularizer
weight $\beta = 7120$ is more than twice as strong as this backbone needs and costs both axes of the
trade-off at once. At $s_{\mathrm{lora}} = 0.40$ that run reaches only TA $72.1$, IA $76.4$ and DINO
$0.536$. Retuned to $\beta = 3000$, the same scale gives TA $75.5$, IA $80.1$ and DINO $0.593$.
Nothing else changes, and the SD-1.5 recipe reported everywhere else in this paper is untouched.

\begin{figure}[h]
  \centering
  \includegraphics{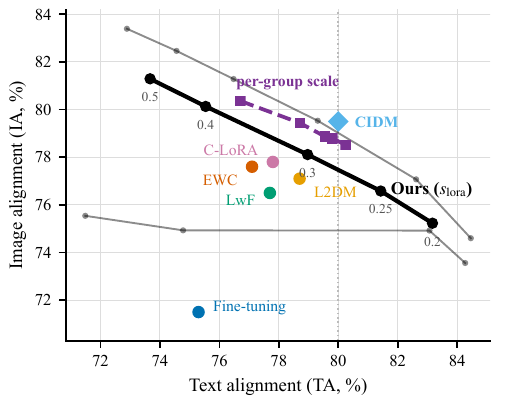}
  \caption{IA versus TA on SDXL for ten concepts. Baseline results are
  taken from the SDXL rows of the CIFC benchmark~\citep{dong2024continually}, and the dotted line
  marks the TA of CIDM. The thick curve shows our method with a single LoRA scale for the
  first training seed, and the two thin curves show two further seeds. The dashed curve keeps the
  scale of the output projection $\mathrm{to\_out}.0$ at $0.50$ while sweeping the other three
  projections, an inference setting of the same checkpoint. All results are for the final model with
  box-free prompts.}
  \label{fig:tradeoff-sdxl}
\end{figure}

\paragraph{Comparison with published results.} Figure~\ref{fig:tradeoff-sdxl} places this run
against the published SDXL rows. Each read at its own TA, the three-seed mean is ahead
of five of the six baselines published in both tables: by $0.3$ points of IA over
C-LoRA, $0.6$ over L2DM, $0.8$ over EWC, $1.6$ over LwF and $7.6$ over sequential fine-tuning, with
the first three inside the seed-to-seed spread. CIDM stays ahead. At its TA of $80.0$,
our IA is $77.5$ on the first seed and $77.1 \pm 2.1$ over three seeds, against $79.5$ for CIDM, a gap of $2.4$ points.

\paragraph{Limitations on this backbone.} Three caveats bound this result. First, the thick curve
in Figure~\ref{fig:tradeoff-sdxl} and the per-group variant below come from a single training seed.
Of two further seeds, one lies above it and one below: at CIDM's TA, they give image
alignments of $79.0$ and $74.9$, against $77.5$. The low seed shows no trade-off. Beyond
$s_{\mathrm{lora}} = 0.20$, raising its scale costs TA without adding IA,
even with its weakest concept left out. We found no cause in the training setup or the data, and we report this backbone as the mean of three seeds without claiming that the recipe is stable on it. Second, this backbone is evaluated on the final model only, without a $55$-cell matrix, and forgetting is measured on SD-1.5 only. Third, the search behind $\beta = 3000$
covered $300$ to $7120$ on this backbone, while every other axis we swept made results worse:
doubling the training budget to $1600$ steps, raising the learning rate to $3 \times 10^{-4}$,
narrowing the task embedding, and dropping the adapter on $\mathrm{to\_out}.0$. The first of these
cost $8$ points of TA at every scale.

\paragraph{A separate scale for the output projection.} The curves above, like every number in this
paper, apply a single scalar $s_{\mathrm{lora}}$ to all four adapted projections, but the four can be
scaled separately. Giving the output projection $\mathrm{to\_out}.0$ its own scalar, held at $0.50$
while the three projections that read the caption are swept, moves the whole front upward. At CIDM's
TA of $80.0$, our IA is $78.7$ instead of $77.5$ and our DINO $0.596$
instead of $0.545$, both interpolated between measured points on each curve. The per-group variant
lies above the single-scalar curve at all five of its measured points, by $0.92$ to $1.21$ points of
IA, and the gap to CIDM's row shrinks from $2.0$ points to $0.8$. This is less than the seed-to-seed spread above, and we do not claim to have closed the gap.

The asymmetry shows where identity lives. If $\mathrm{to\_out}.0$ is instead pinned low, at $0.12$,
sweeping the three reading projections across $4.9$ points of TA moves IA
by $0.46$ and DINO not at all ($0.475$ to $0.479$), where the single scalar moves IA by
$3.1$. The projections that read the caption contribute almost no identity on their own, and the
output projection gates it. The ablation in Appendix~\ref{app:ablations} reaches the same conclusion
by removing that projection, whereas here it follows without removing anything. The SD-1.5 numbers
keep the single scale.

\section{Ablations}
\label{app:ablations}

Table~\ref{tab:ablations} lists every ablation of the recipe on the ten CIFC concepts. Each
variant changes a single component, is trained with the seed of the reference row (unless marked
$^{\dagger}$), and is evaluated on its own full $55$-cell matrix at $s_{\mathrm{lora}} = 0.45$. The
two architecture and data rows are evaluated on the final model only.

\begin{table}[h]
  \centering
  \caption{Ablations of the recipe, ten concepts, SD-1.5, one training seed per variant (the seed of the reference row), full $55$-cell matrix at $s_{\mathrm{lora}} = 0.45$. TA and IA in percent, DINO of the final model, forgetting as earlier peak minus final DINO (Appendix~\ref{app:metrics}). We claim no difference below $0.0091$ DINO. The reference row is one seed of the reported model, whose three-seed values are $0.617$ and $0.0060$. Rows marked $^{\dagger}$ are means over the same three training seeds.}
  \label{tab:ablations}
  \footnotesize
  \begin{tabular}{llcccc}
    \toprule
    & Variant & TA $\uparrow$ & IA $\uparrow$ & DINO $\uparrow$ & Forgetting $\downarrow$ \\
    \midrule
    & reference ($\beta = 7120$, lookahead, penalty on $\Delta\mathbf{W}$) & $75.27$ & $79.46$ & $0.621$ & $0.0094$ \\
    \midrule
    \multicolumn{6}{l}{\emph{Regularizer}} \\
    & $\beta = 0$ & $76.88$ & $78.39$ & $0.589$ & $0.0213$ \\
    & $\beta / 4$ & $75.35$ & $79.52$ & $0.620$ & $0.0018$ \\
    & $4 \beta$ & $76.42$ & $78.43$ & $0.612$ & $0.0077$ \\
    & no lookahead, $\beta$ & $75.32$ & $79.63$ & $0.624$ & $0.0071$ \\
    & no lookahead, $4 \beta$ & $75.59$ & $79.06$ & $0.615$ & $0.0088$ \\
    & penalty on the factors, $\beta = 70$ & $75.10$ & $79.57$ & $0.623$ & $0.0047$ \\
    & penalty on the factors, $\beta = 700$ & $75.53$ & $79.40$ & $0.624$ & $0.0065$ \\
    & penalty on the factors, $\beta = 7120$ & $75.27$ & $79.66$ & $0.628$ & $-0.0038$ \\
    \midrule
    \multicolumn{6}{l}{\emph{Task embedding}} \\
    & no Gram--Schmidt$^{\dagger}$ & $75.89$ & $79.20$ & $0.627$ & $0.0030$ \\
    & task embedding from the reference images$^{\dagger}$ & $74.28$ & $79.85$ & $0.618$ & $0.0163$ \\
    & learned per-concept modulation & $76.67$ & $78.71$ & $0.615$ & $0.0024$ \\
    \midrule
    \multicolumn{6}{l}{\emph{Architecture and data}} \\
    & without $\mathrm{to\_out}.0$ & $68.7$ & $68.9$ & $0.347$ & -- \\
    & without composites & $78.17$ & $77.37$ & $0.580$ & -- \\
    \midrule
    \multicolumn{6}{l}{\emph{Concept order}} \\
    & random permutation 1 & $76.46$ & $78.46$ & $0.616$ & $0.0031$ \\
    & random permutation 2 & $74.61$ & $79.40$ & $0.626$ & $0.0095$ \\
    \bottomrule
  \end{tabular}
\end{table}

Among the regularizer and embedding variants, only removing the regularizer costs more than the
seed-to-seed threshold of $0.0091$ DINO. Tying the task embedding to the concept's appearance
preserves identity over three seeds but nearly triples forgetting. The lookahead step of
Eq.~\ref{eq:anchor} and placing the penalty on $\Delta\mathbf{W}$ rather than on its factors show no
measurable difference, and $\beta$ can be scaled by a factor of four in either direction without
effect. Removing Gram--Schmidt has no measurable cost on these metrics over three seeds either. Its
effect shows in how far old updates drift, which Appendix~\ref{app:embeddings} measures in weight
space. The order of the sequence does not matter: two random permutations of the ten concepts change
DINO by at most $0.005$ and forgetting by at most $0.0063$, both within the threshold.

For two rows, DINO understates the effect. Without composites, placement is no longer learned. IoU
with the requested box falls from $0.766$ to $0.270$, and the quadrant is correct in $62\%$ of
generations against $100\%$ for the reference. The photographs alone do not teach placement
(\S\ref{sec:localization}). Without $\mathrm{to\_out}.0$, the adapter loses most of the concept at
every scale, as the next paragraph shows.

\paragraph{Adapting the output projection.} Which layers are adapted is not merely a
question of parameter budget. Retraining the reported recipe with $\mathrm{to\_out}.0$ removed from
the adapted set, everything else unchanged, costs far more than the quarter of the parameters it
saves. After ten tasks at $s_{\mathrm{lora}} = 0.45$, DINO falls from $0.621$ to $0.347$, image
alignment from $79.5$ to $68.9$, and TA from $75.3$ to $68.7$. Raising the scale does
not recover the loss ($0.379$ at $0.75$). One concept, the dog, collapses entirely, but the
remaining nine alone reach $0.389$ at $0.45$, which shows that the drop is not caused by a single outlier. The
same holds after fifty tasks.

The ablation is equally costly on SDXL-base-1.0. DINO falls from $0.593$ to $0.430$ and image
alignment from $80.1$ to $72.1$, with the ablated variant evaluated at the best of three adapter
scales, which shows that the effect comes from the construction rather than from the backbone it was designed on.
TA is the only metric that does not drop there ($75.5$ to $76.0$). This is what one
would expect from removing a projection that writes into the residual stream: the model still
follows the caption, but less of the concept appears in the image.

We emphasize this point because Custom Diffusion~\citep{kumari2023customdiffusion}, a common
per-concept adapter, adapts only $\mathrm{to\_k}$ and $\mathrm{to\_v}$. A per-concept adapter can
afford this because it is fitted directly. A hypernetwork that must \emph{generate} the adapter does
not have that freedom. The two projections that read the caption are token-masked and act on only a
handful of positions, so without $\mathrm{to\_out}.0$ the generated update has too little of the
residual stream left to write into.

\section{Concepts along the sequence}
\label{app:forgetstream}

\begin{figure}[h]
  \centering
  \includegraphics{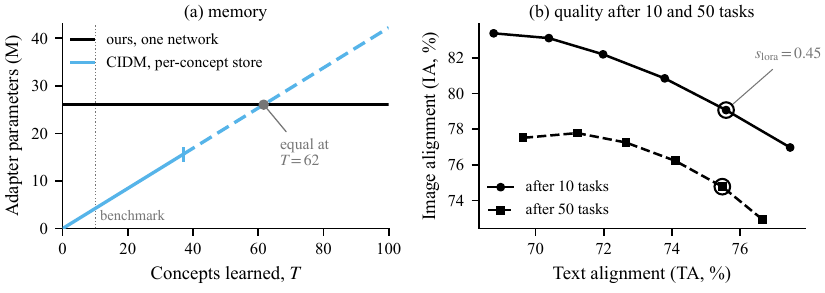}
    \caption{Scaling beyond the ten concepts of CIFC. \textbf{(a)} Stored parameters of our
  fixed-size network and of CIDM, which stores an adapter and its token embeddings for every concept
  ($0.422$M, our measurement). The two are equal at $T = 62$, and the CIDM line is dashed beyond
  $37$ concepts, where its implementation breaks down. \textbf{(b)} IA versus text
  alignment of the first ten concepts after ten and after fifty tasks, obtained by varying the LoRA
  scale. After fifty tasks the curve lies lower at nearly the same TA, so these concepts
  lose part of their identity. Rings mark the reported scale, and results are means over three
  training seeds with box-free prompts.}
  \label{fig:scaling}
\end{figure}

\begin{figure}[h]
  \centering
  \input{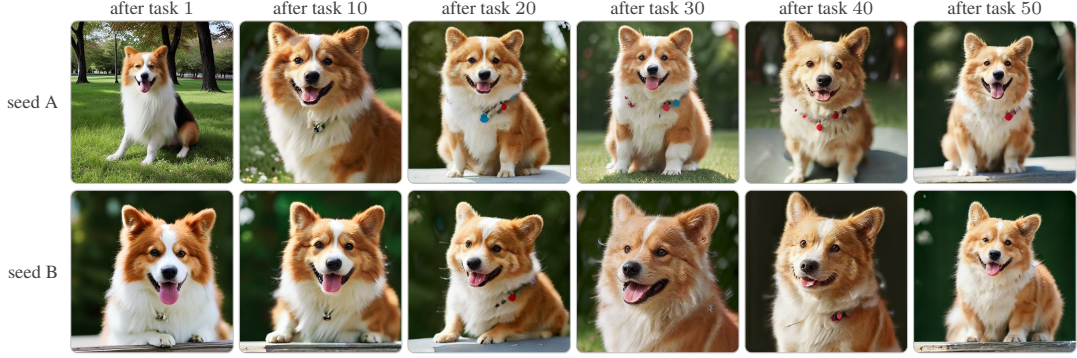}
  \caption{Retention of the first concept along the fifty-concept sequence. The columns show
  generations from the checkpoints saved after $1$, $10$, $20$, $30$, $40$ and $50$ tasks at
  $s_{\mathrm{lora}} = 0.45$. Each row uses one seed with the prompt and initial noise fixed, so every
  change reflects the concepts learned in the meantime. The markings and the face of the concept are
  preserved throughout, while the framing and the props added by the scene vary.}
  \label{fig:forgetgrid}
\end{figure}

\begin{figure}[h]
  \centering
  \input{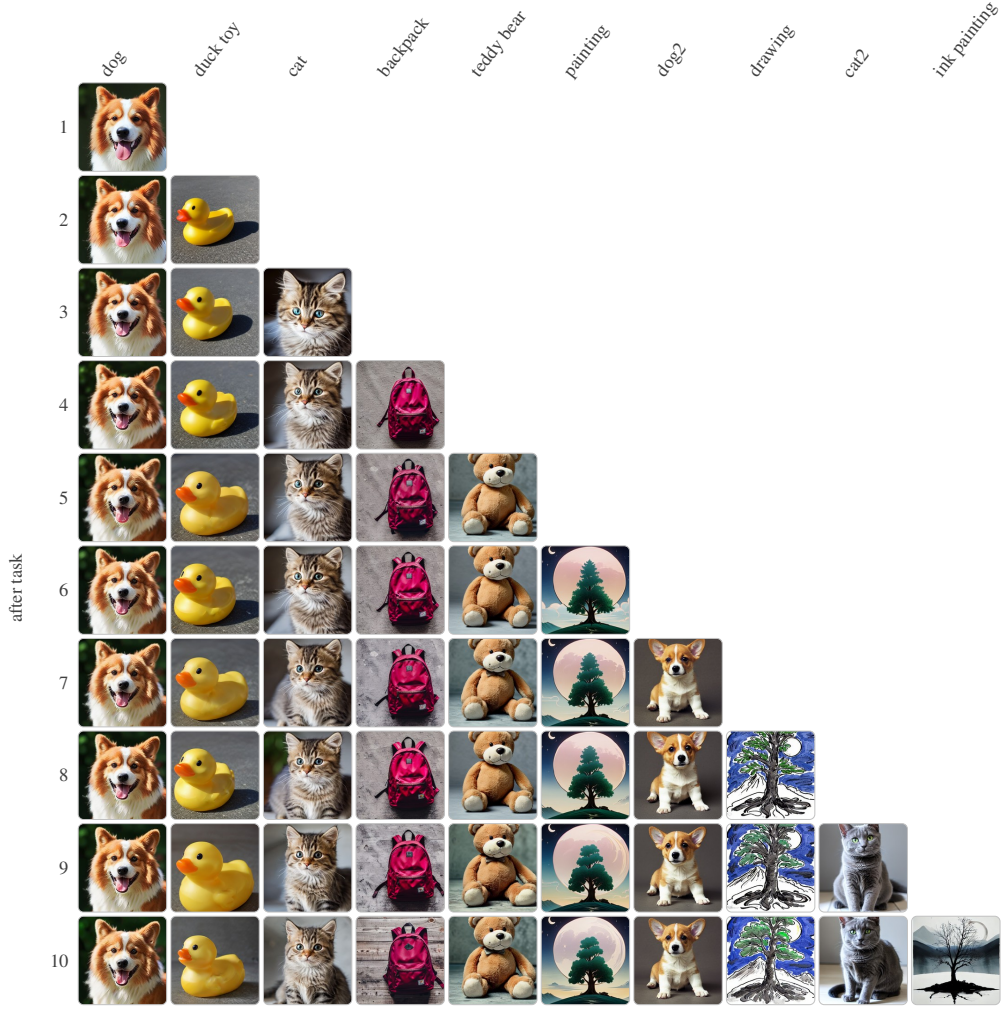}
  \caption{Retention of the ten CIFC concepts along the ten-concept sequence, for one training
  seed. Row $k$ shows the model saved after task $k$ and column $j$ shows concept $j$, so a cell
  appears once its concept has been learned. The prompt and initial noise are fixed within each
  column ($s_{\mathrm{lora}} = 0.45$), so changes within a column reflect the later tasks. Our method
  preserves every concept: the markings of the dogs and cats, the crimson of the backpack and the
  shape of the duck toy remain throughout, although the prompt never names a color, while the framing
  and the background vary.}
  \label{fig:forget10}
\end{figure}

Figure~\ref{fig:scaling} summarizes the longer sequences of \S\ref{sec:results}: (a) the storage of the
fixed-size network against a store that grows with every concept, and (b) the ten CIFC concepts after ten and after fifty tasks as curves over the LoRA scale. Figures~\ref{fig:forgetgrid}
and~\ref{fig:forget10} show the same sequences as images, the first concept over fifty tasks and all
ten concepts over ten. Read at full adapter strength instead of $s_{\mathrm{lora}} = 0.45$, the first
concept desaturates and develops texture artifacts from the tenth task on, and its DINO falls by
$0.16$ by the fiftieth. Of four further concepts measured the same way, three score \emph{higher} at
full strength by the fiftieth task. The scale trades identity against prompt fidelity
differently for each concept, and the single scale we report is a compromise across them.

\section{Analysis of the task embeddings}
\label{app:embeddings}

This appendix examines the task embeddings: what tying them to appearance costs, how the updates
they generate relate to one another, how far those updates drift along the sequence, and what the
network generates away from the embeddings it was trained on.

\paragraph{Task embeddings tied to appearance.} One ablation replaces half of the initial task embedding, $128$ of its $256$
dimensions, with a fixed projection of the mean CLIP \emph{image} embedding of the concept's
reference photos, and keeps the rest of the recipe. Scored on its full $55$-cell matrix at
$s_{\mathrm{lora}} = 0.45$, it leaves identity unchanged over three training seeds ($0.618$ against $0.617$) but raises forgetting
to $0.016$, close to three times the reported value. Tying the task embedding to the concept's appearance
therefore gains nothing and costs stability, which is why we keep the task embedding arbitrary.

\begin{figure}[h]
  \centering
  \includegraphics{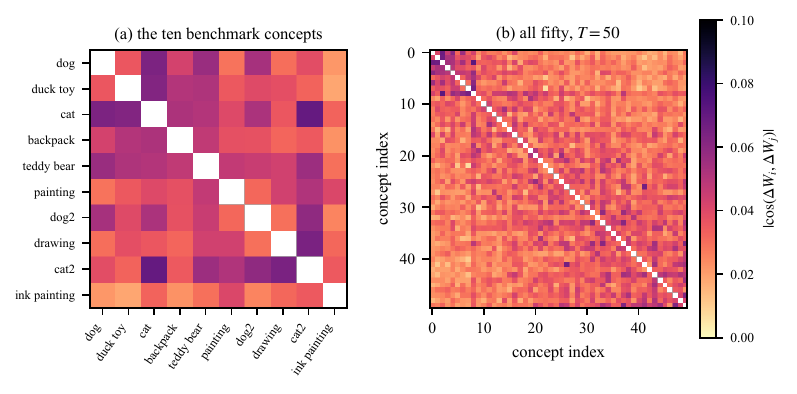}
  \caption{Cosine similarity between the generated updates of every pair of concepts,
  summarized by a single number in Table~\ref{tab:ortho}. \textbf{(a)} The ten CIFC concepts. The
  largest value is between cat and cat2 ($0.070$), followed by drawing and cat2, dog and cat, and duck
  toy and cat ($0.062$ to $0.064$), so the alignment is small everywhere and only partly semantic.
  \textbf{(b)} The same after fifty tasks, where the block of the ten CIFC concepts is slightly darker
  than the remaining forty and no other structure appears. The diagonal is masked. The cosine is
  averaged over layers before taking the absolute value, whereas Table~\ref{tab:ortho} takes the
  absolute value per layer, so the values here are smaller ($0.042$ against $0.056$ off the diagonal
  at $T = 10$). Signs differ across layers, which indicates that the alignment is not a single shared
  direction.}
  \label{fig:orthoheat}
\end{figure}

\paragraph{Orthogonality of the generated updates.} We take the trained model and
compute, for every pair of concepts and every adapted layer, the cosine between the generated
updates, $\langle \Delta\mathbf{W}_{\phi,\ell}(\mathbf{v}_i), \Delta\mathbf{W}_{\phi,\ell}(\mathbf{v}_j) \rangle_F /
(\lVert \Delta\mathbf{W}_{\phi,\ell}(\mathbf{v}_i) \rVert_F \lVert \Delta\mathbf{W}_{\phi,\ell}(\mathbf{v}_j) \rVert_F)$, using the same $r \times r$ Gram matrices as in Appendix~\ref{app:anchor},
so that no $\Delta\mathbf{W}$ is ever formed. Table~\ref{tab:ortho} reports the mean absolute off-diagonal
cosine. The updates of different concepts sit at $0.056$ after ten tasks and $0.044$ after fifty,
compared with $0.27$ for the $\mathbf{A}$ factors that produce them and $0.001$ for random
matrices of the same shape. The closest pairs are the two cats ($0.07$) and a cat with a dog
($0.06$), and Figure~\ref{fig:orthoheat} shows that no pair stands far above the rest. Nothing in
the objective asks for this, and neither does the geometry of the inputs: retraining without
Gram--Schmidt, that is, with random non-orthogonal embeddings, reproduces the same overall figures
($0.056$ at $T = 10$ and $0.043$ at $T = 50$, against $0.056$ and $0.044$ here), with every
projection within $0.006$. Near-orthogonality of the generated updates therefore comes from the generator itself, and the
orthogonalized inputs do not create it.

Gram--Schmidt leaves this statistic unchanged because of the regime we operate in. The embeddings
are $d_v = 256$ wide, and two random unit vectors in that many dimensions already meet at an
expected cosine of $\sqrt{2 / (\pi d_v)} \approx 0.05$. At $T = 50$ the space is far from crowded,
and forcing the task embeddings apart changes little about how the updates sit relative to one another. The
change from $T = 10$ to $T = 50$ agrees: the mean cosine falls rather than rises as concepts are added. More concepts do not crowd the output space. A random task embedding overlaps more with the span of
the earlier ones as $T$ approaches $d_v$, whereas Gram--Schmidt keeps the task embeddings exactly orthogonal up
to $T = d_v = 256$ at no cost in parameters. What Gram--Schmidt does change is how far each update
moves, which the next paragraph measures.

\begin{table}[h]
  \centering
  \caption{Mean absolute cosine between the generated updates of different concepts, per adapted
  projection. Factors are the same quantity computed on $\mathbf{A}$ and $\mathbf{B}$ separately. Random is the same statistic on matrices of the same shape with i.i.d.\ entries, quoted at $T = 10$. The same reference at $T = 50$ differs by at most $0.0002$.}
  \label{tab:ortho}
  \footnotesize
  \begin{tabular}{lccccccc}
    \toprule
    & \multicolumn{2}{c}{$\Delta\mathbf{W}$} & \multicolumn{2}{c}{factor $\mathbf{A}$} & \multicolumn{2}{c}{factor $\mathbf{B}$} & random \\
    \cmidrule(lr){2-3} \cmidrule(lr){4-5} \cmidrule(lr){6-7}
    Projection & $T = 10$ & $T = 50$ & $T = 10$ & $T = 50$ & $T = 10$ & $T = 50$ & \\
    \midrule
    $\mathrm{to\_q}$   & $0.039$ & $0.035$ & $0.16$ & $0.17$ & $0.24$ & $0.19$ & $0.0013$ \\
    $\mathrm{to\_k}$   & $0.070$ & $0.060$ & $0.34$ & $0.33$ & $0.20$ & $0.18$ & $0.0011$ \\
    $\mathrm{to\_v}$   & $0.088$ & $0.046$ & $0.36$ & $0.30$ & $0.24$ & $0.15$ & $0.0011$ \\
    $\mathrm{to\_out}$ & $0.026$ & $0.037$ & $0.21$ & $0.27$ & $0.11$ & $0.12$ & $0.0014$ \\
    \midrule
    all                & $\mathbf{0.056}$ & $\mathbf{0.044}$ & $0.27$ & $0.27$ & $0.20$ & $0.16$ & $0.0012$ \\
    \bottomrule
  \end{tabular}
\end{table}

\begin{figure}[h]
  \centering
  \includegraphics{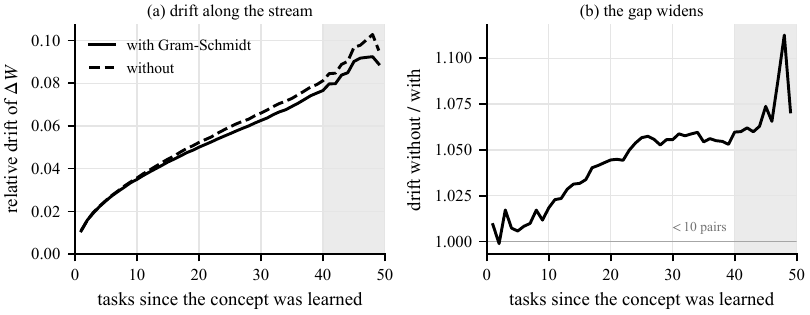}
  \caption{Drift of the generated update along the fifty-concept sequence
  (Eq.~\ref{eq:drift}), averaged over all concept--checkpoint pairs at each lag. \textbf{(a)} Drift
  with and without Gram--Schmidt orthogonalization of the task embeddings. \textbf{(b)} The ratio of
  the two, which is close to one at short lags and grows with the length of the sequence. Beyond a lag
  of ten, the average at lag $L$ covers $50 - L$ pairs, and the shaded region, with fewer than ten
  pairs, reflects sampling noise. All values are computed from checkpoints, without generating
  images.}
  \label{fig:drift}
\end{figure}

\paragraph{Drift of the generated updates.} The regularizer asks that
the update generated for an old embedding stay where it was, and we measure how far it moves. For
concept $j$ and any later checkpoint $k \ge j$, we compute
\begin{equation}
  d(j, k) \;=\; \frac{\lVert \Delta\mathbf{W}_{\phi_k,\ell}(\mathbf{v}_j) - \Delta\mathbf{W}_{\phi_j,\ell}(\mathbf{v}_j) \rVert_F}
                     {\lVert \Delta\mathbf{W}_{\phi_j,\ell}(\mathbf{v}_j) \rVert_F},
  \label{eq:drift}
\end{equation}
the relative movement of that concept's own update, and average it over all pairs with the same
lag $k - j$. Expanding Eq.~\ref{eq:drift} requires three $r \times r$ Gram matrices per layer (the
identity of Appendix~\ref{app:anchor}, extended to two sets of factors). With it, the full fifty-task sweep reads only checkpoints, generates no images and takes about a minute.
Figure~\ref{fig:drift} shows the result. On average, a generated update moves by $1.1\%$ of its
own norm after one further task, and the first concept's update moves by $8.9\%$ after
forty-nine. The additional drift per task decreases over the first twenty tasks and stays near
$0.13\%$ per task after that.

The measurement is stable across training runs. An independently retrained run of the same
fifty-concept sequence reproduces the curve to within $4\%$ at every lag: $1.1\%$ after one further
task, as here, and $8.7\%$ after forty-nine, against $8.9\%$. At every lag, the two runs average
over the same set of pairs, so only the network differs.

On this measurement, the two embedding geometries do separate. At a lag of one task, the
orthogonalized and non-orthogonalized runs are indistinguishable, with a ratio of $1.009$. The gap
widens to $1.04$ at a lag of twenty and $1.07$ at forty-nine. Gram--Schmidt therefore does add
stability to the generated update, but the effect is small and accumulates with the length of the
sequence rather than being present from the start. The longest lags average over few pairs, only one
at a lag of forty-nine, which is why the figure shades them.

\begin{figure}[h]
  \centering
  \input{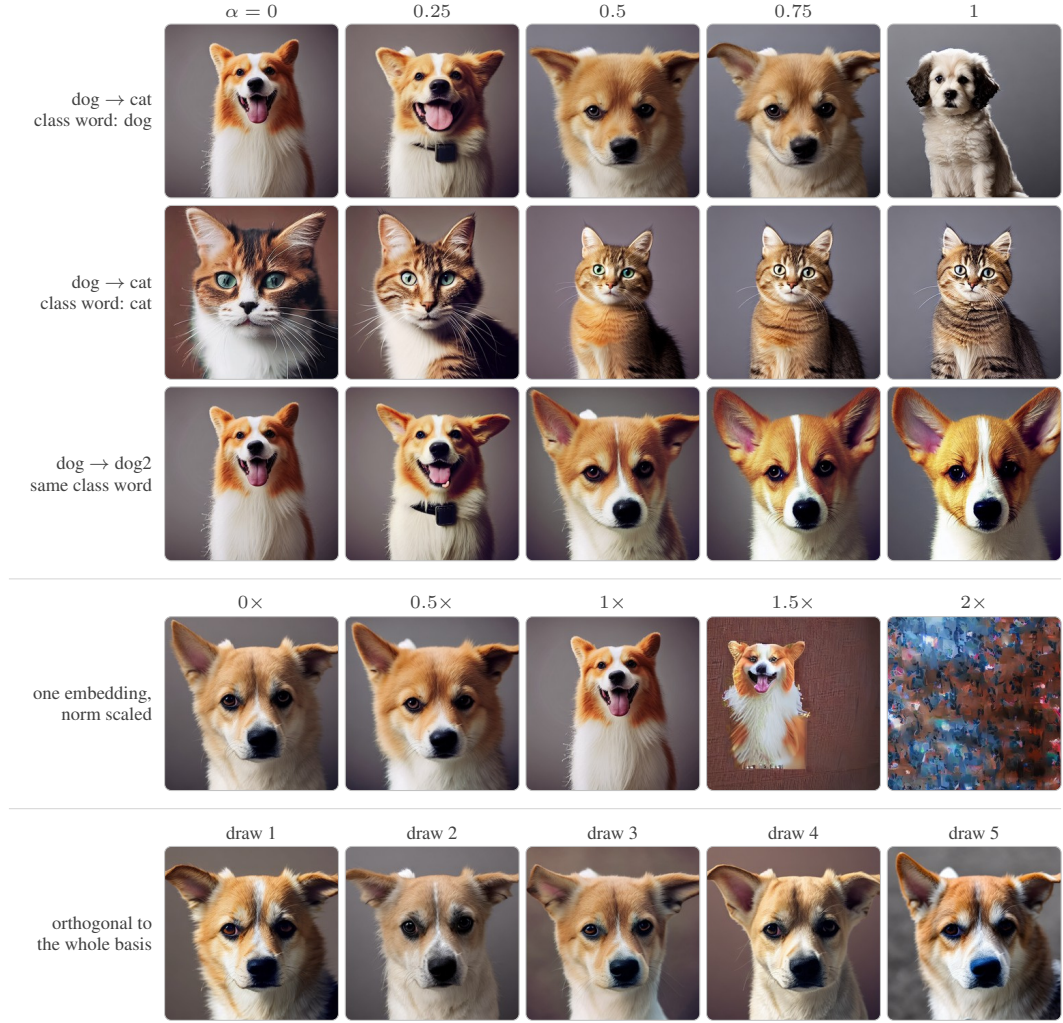}
  \caption{Generations between and beyond the learned task embeddings, with the placement
  branch disabled. \textbf{Rows 1--2:} Interpolation from $\mathbf{v}_{\mathrm{dog}}$ to
  $\mathbf{v}_{\mathrm{cat}}$, rendered with each of the two class words. The category follows the
  class word and the identity follows the embedding, so the far end is a generic member of the
  requested class and no hybrid appears. \textbf{Row 3:} Interpolation between two concepts of the
  same class, where the identity changes continuously. \textbf{Row 4:} One embedding scaled from $0$
  to $2\times$ its norm, which acts like the LoRA scale and breaks down beyond $1.5\times$.
  \textbf{Row 5:} Five random vectors orthogonal to all learned embeddings, each producing a clean,
  generic instance of the class. Each cell uses one seed, and a second seed gives the same result.}
  \label{fig:embedsheets}
\end{figure}

\paragraph{Probing the embedding space.} During training, the hypernetwork only
ever sees the task embeddings themselves, so we ask what it generates elsewhere in the embedding
space (Figure~\ref{fig:embedsheets}). Generating along the line between two embeddings gives a
smooth transition. Between two dogs, the identity morphs continuously rather than switching (third
row). Between a dog and a cat, the class word determines the category and the embedding determines the identity within it. No point on the line mixes the two concepts, and the far end is a generic animal of
the class the prompt asks for (first two rows, the same interpolation rendered once with each class
word). Scaling one embedding acts like an adapter strength. At zero and at half the norm the model
returns the class prior, and at the full norm the concept. At $1.5\times$ the image starts to break
up, and at twice the norm it is noise. Five random vectors orthogonal to the whole basis, scaled to
the median norm of the task embeddings, each produce a clean but generic instance of the class word, which shows that an unused embedding does no harm. These are inference-only probes, run with the localization branch off at
$s_{\mathrm{lora}} = 0.6$.

\section{Evaluating the placement module}
\label{app:grids}

\subsection{Following the requested box}

Table~\ref{tab:grounding} evaluates placement under its own protocol: scene prompts, one quadrant
box per generation, and $84$ generations per configuration. TA and identity are
therefore comparable down the columns of the table but not with Figure~\ref{fig:tradeoff}. Identity
is reported on the subject's mask. The reference images are close-ups, so a subject correctly
confined to a quarter of the canvas is rendered at half the linear resolution and scores lower on
the whole image regardless of its quality.

\begin{table}[h]
  \centering
  \caption{Placement on the seven object concepts, quadrant boxes, scene prompts, $84$ generations per row, one checkpoint. \emph{Quadrant}: the quadrant crop that best matches the concept's references is the given one (no detector, chance $25\%$). DINO$_{\mathrm{mask}}$: identity on the subject's detected mask. The upper block is what the method is measured against, and the last row is the setting we report. Sampling noise on IoU $>0.5$ is up to $10$ points. The other columns repeat across seed batches. Designs we measured and did not adopt are in Appendix~\ref{sec:negative}.}
  \label{tab:grounding}
  \footnotesize\setlength{\tabcolsep}{2pt}
  \begin{tabular}{lccccccc}
    \toprule
    Configuration & Quadrant $\uparrow$ & IoU $>0.5$ & Contain.\ $\uparrow$ & Fill $\rightarrow 1$ & TA $\uparrow$ & DINO$_{\mathrm{mask}}$ $\uparrow$ & Det. \\
    \midrule
    No box                                   & $25\%$ & $0\%$  & $0.25$ & $1.07$ & $0.778$ & $0.599$ & $72/84$ \\
    Position named in the prompt             & $21\%$ & $0\%$  & $0.25$ & $1.01$ & $0.743$ & $0.597$ & $67/84$ \\
    Training-free layout, same adapters      & $49\%$ & $2\%$  & $0.37$ & $0.92$ & $0.770$ & $0.528$ & $65/84$ \\
    \midrule
    Placement module, $\kappa = 1$        & $93\%$ & $52\%$ & $0.74$ & $1.34$ & $0.780$ & $0.713$ & $84/84$ \\
    Placement module, $\kappa = 2$        & $\mathbf{100\%}$ & $\mathbf{96\%}$ & $\mathbf{0.93}$ & $1.07$ & $0.795$ & $\mathbf{0.724}$ & $82/84$ \\
    $\kappa = 2$, $+$ bootstrapping          & $\mathbf{100\%}$ & $\mathbf{96\%}$ & $\mathbf{0.93}$ & $\mathbf{1.01}$ & $\mathbf{0.828}$ & $0.717$ & $84/84$ \\
    \bottomrule
  \end{tabular}
\end{table}

\paragraph{Contribution of each component.} The gain provides the placement: going from
$\kappa = 1$ to $\kappa = 2$ nearly doubles the share of generations with IoU above $0.5$.
Bootstrapping leaves IoU and containment unchanged but brings the fill ratio to $1.01$, recovers the
two generations in which the branch alone lost the subject, and restores the TA that
the gain costs ($0.828$ against $0.795$, and above the no-box row at $0.778$), because the background
outside the box is painted from the same prompt.

\paragraph{Boxes other than quadrants.} We repeat the test with nine small boxes on a $3 \times 3$
grid, each a third of the side, and with four elongated strips ($0.35 \times 0.9$ and
$0.9 \times 0.35$), after ten tasks and on three training seeds. The concept lands in the requested
cell in $86\%$ of generations (chance $11\%$) and in the requested strip in $94\%$ (chance $25\%$).
Placement is less precise than with quadrants. IoU is $0.565$ and $0.570$, against $0.71$, and
$74\%$ and $64\%$ of detected generations exceed IoU $0.5$. A small box is slightly overfilled
(fill ratio $1.15$), and a strip receives a subject of the right area that does not always take its
shape (containment $0.83$). Placement therefore follows boxes of other shapes and sizes, but matches
their extent best at the size of a quadrant.

\begin{figure}[h]
  \centering
  \input{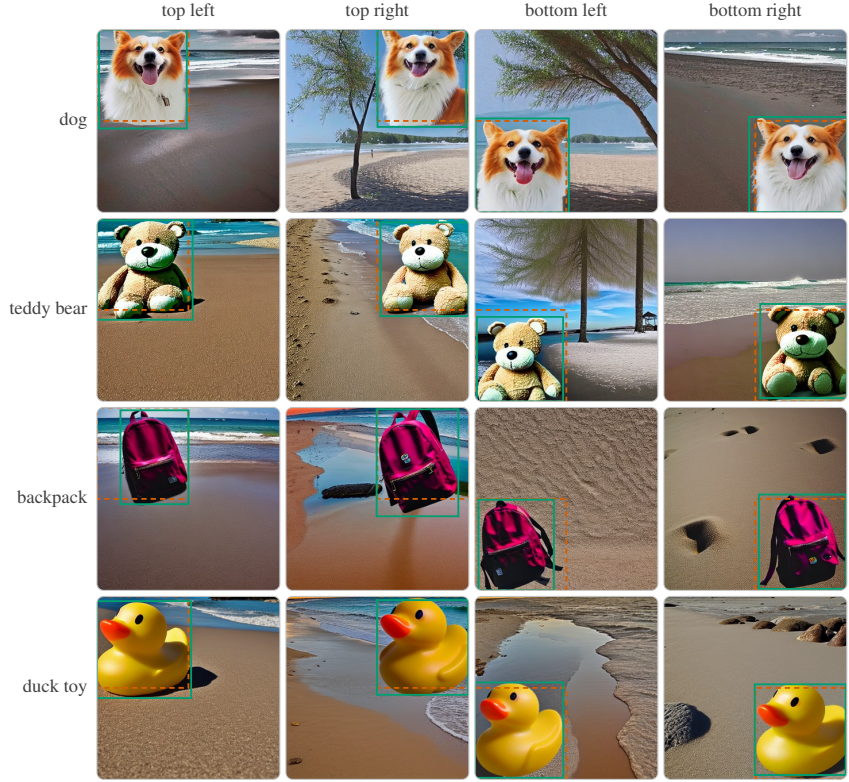}
  \caption{Placement of the four object concepts in the four quadrant boxes, at the reported
  setting ($\kappa = 2$, bootstrapping over the first $K = 15$ of $50$ steps,
  $s_{\mathrm{lora}} = 0.7$). Rows show the dog, teddy bear, backpack and duck toy, and columns show
  the requested quadrants. The dashed box is the requested one and the solid box is the one detected
  by a Mask R-CNN, and the two coincide so closely that the dashed line is visible only where it
  extends beyond the solid one. The scene prompt is the same in every panel, so the background varies
  only with the seed and the box.}
  \label{fig:grids}
\end{figure}

\paragraph{Qualitative check.} As Appendix~\ref{sec:pitfalls} argues, the placement numbers must be
read together with images, because containment also rises when part of the subject is removed and a
duplicated subject still yields a well-formed detection box. Figure~\ref{fig:grids} shows the grid at
the reported setting: four concepts and the four quadrant boxes, one sample each, with the requested
box dashed and the detected box solid.

\subsection{Placement along the sequence}
\label{sec:transfer}

The placement module keeps training on every task, so the placement of an early concept can
change after its own task has ended. We test the models saved after tasks $1$, $5$, $10$ and $50$
under the protocol of Table~\ref{tab:grounding}, on three training seeds.

Over the first ten tasks, placement transfers backward. The first concept is placed better once
later concepts are learned: its IoU rises from $0.60$ after its own task to $0.75$ after task $10$,
on every seed. Over fifty tasks this gain reverses. IoU on the seven objects falls from $0.71$ after
ten tasks to $0.55$ after fifty, and the subject is rendered smaller than its box. It still lands in
the requested quadrant in most generations ($73\%$).

\paragraph{Regularizing the placement tokens.} The output-space regularizer protects the generated adapters
but not the placement tokens. We therefore extended Eq.~\ref{eq:anchor} to the tokens generated for
earlier embeddings and, in a second variant, also to the per-layer gates. Neither variant preserves
placement. Both lower it from the first task on, and both weaken the content module: after ten
tasks DINO falls from $0.621$ to $0.493$ and $0.550$. We therefore leave the tokens unregularized. These
runs use one training seed each and predate the final segmentation masks, so their numbers are
comparable only with one another.

\subsection{Rejected alternatives to the placement module}
\label{sec:negative}

Each variant below was trained and tested at the point in development where it was proposed, so each
number compares the variant with the design current at that time, not with the final checkpoint.
Table~\ref{tab:alternatives} lists six designs lighter than Eq.~\ref{eq:gsa}, evaluated on a
left/right test: a box covering the left or the right half of the frame, $84$ samples, and a pass
threshold of $80\%$ fixed before the tests. The gate magnitudes are comparable across all six, so the
difference lies in the capacity of the branch.

\begin{table}[h]
  \centering
  \caption{Left/right placement test for the designs we measured and did not adopt. Percentage of generations in which the subject is detected on the requested side. The pass threshold was fixed at $80\%$ before the tests. The last row is the attention branch as it stood when the six were measured, not the reported model, whose placement is in Table~\ref{tab:grounding}.}
  \label{tab:alternatives}
  \begin{tabular}{lc}
    \toprule
    Design & Left/right $\uparrow$ \\
    \midrule
    Box mapped into the adapter weights & $48.8\%$ \\
    Scalar sigmoid gate on cross-attention & $48.8\%$ \\
    Five times more paste augmentation & $47.6\%$ \\
    Learned geometry, layout-conditioned style & $51.2\%$ \\
    Segmentation-based objective alone & $50.0\%$ \\
    Box mask with a scalar gate & $47.6\%$ \\
    \midrule
    Attention branch, Eq.~\ref{eq:gsa} & $\mathbf{84.5\%}$ \\
    \bottomrule
  \end{tabular}
\end{table}

Four further mechanisms were measured and rejected after the main design was fixed. Restricting the
injection to coarse attention resolutions costs $16$ points of IoU $>0.5$ and improves color by only
$0.007$, so the finest layers take part in the layout. A penalty on the cross-attention logits of the
concept's tokens, applied at every image position outside the box, has no effect: $62\%$, $64\%$,
$64\%$ and $65\%$ of generations reach IoU $>0.5$ at penalties of $0$, $3$, $6$ and $10$. The text
encoder is causal, so every token after the concept still carries it. Extending the penalty to the
whole causal tail raises containment to $0.81$, against $0.67$ for the branch alone, but in the one
pair where TA was also scored it costs six points ($80.7$ to $74.6$). The words that
describe the scene come after the concept in the prompt and are suppressed along with it, which is
why the method adds content inside the box and suppresses nothing. Isolating self-attention inside
the box destroys placement (from $75\%$ to $4\%$), which shows that the region outside the box
generates the subject independently.

Feathering the bootstrapping mask reduces detection to $74$ of $84$ generations and desaturates the
outputs, for the reason given in Appendix~\ref{app:details}. The background that bootstrapping paints
outside the box could instead be drawn from a set of images rendered in advance. On this benchmark,
such a background contains the subject no better than the prompt-derived one ($0.92$ in both) and
places it worse (IoU $0.63$ against $0.73$, and $76\%$ against $90\%$ of generations above $0.5$),
because the scene it paints disagrees with the one the prompt asks for. It also requires that set of
images at inference.

The placement controls leave the no-box protocol of Table~\ref{tab:grounding} unchanged. With every
control enabled and no box supplied, identity, color, fill and IoU stay within sampling noise of a
clean run, so the results with and without a box come from one checkpoint under one protocol.

\section{Multi-concept composition}
\label{sec:composition}

Because CIFC defines no protocol for scoring scenes with several concepts, this appendix reports
our own mechanism on its own terms. We follow the region-based sampling of
CIDM~\citep{dong2024continually}. At every step, each box is denoised with its concept's prompt, the
adapter generated for that concept and the placement module addressed by that box, and the region
predictions are merged with a global one. In CIDM, each region is also helped by the token embeddings
that CIDM learns for its concept at every layer. Our network keeps no per-concept tokens, so its regions
rely on the generated adapter and the placement module alone. We run the eleven composite scenes of CIFC on SDXL, with four seeds each and one setting for all of them.

Inside the boxes, composition works. Of the $124$ regions the scenes request, the requested concept is
the best match for its own region in $121$. The regions also keep the identity of their concepts, with
a mean DINO of $0.716$ against their own references.

\begin{figure}[h]
  \centering
  \input{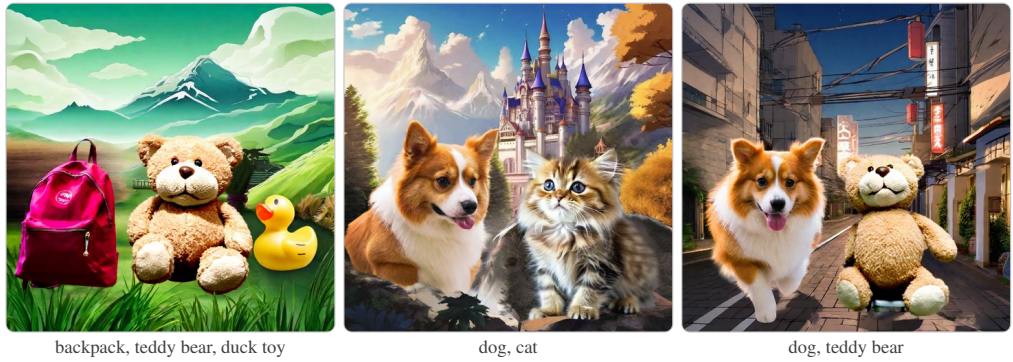}
  \caption{Multi-concept composition on SDXL, three frames generated with one setting at LoRA
  scale $0.55$. Every requested concept is present and recognizable. The scene is shared rather than
  pasted: the grass in the left frame passes in front of all three objects, and no seam follows a box
  boundary. The scenes and boxes are taken from the CIFC benchmark, while the sampler, step count and
  seed are ours.}
  \label{fig:compose}
\end{figure}

The concepts become part of one scene rather than a collage. They share the lighting and drawing
style of the background and sit at a plausible depth within it (Figure~\ref{fig:compose}).

\end{document}